\documentclass{article}

\PassOptionsToPackage{numbers, compress}{natbib}
\usepackage[preprint]{neurips_2024}

\usepackage[utf8]{inputenc} % allow utf-8 input
\usepackage[T1]{fontenc}    % use 8-bit T1 fonts
\usepackage{hyperref}       % hyperlinks
\usepackage{url}            % simple URL typesetting
\usepackage{booktabs}       % professional-quality tables
\usepackage{amsfonts}       % blackboard math symbols
\usepackage{nicefrac}       % compact symbols for 1/2, etc.
\usepackage{microtype}      % microtypography
\usepackage{xcolor}         % colors
\usepackage{lipsum}
\usepackage{amsmath}
\usepackage{amsthm}
\usepackage{nicefrac}  
\usepackage{mathtools}
\usepackage{mathrsfs}
\usepackage{bm}
\usepackage{subfloat}
\usepackage{subfig}
\usepackage{cleveref}
\usepackage[pdftex]{graphicx}   % REMOVE draft if you want pictures
\usepackage{wrapfig}
\usepackage{footmisc}
\usepackage{epstopdf}
\usepackage{svg}
\usepackage[normalem]{ulem}
\useunder{\uline}{\ul}{}
\usepackage{floatflt}
\usepackage{placeins}
\usepackage{tikz-cd}
\usepackage{array}
\usepackage{xcolor}
\usepackage{float}

\title{LaT-PFN: A Joint Embedding Predictive Architecture for In-context Time-series Forecasting}

\author{
  Stijn Verdenius\thanks{Equal contributions} \\ %\thanks{Use footnote for providing further information
  WAIR\thanks{Website: \url{https://wair.ai/}}, Amsterdam \\
  \texttt{stijn@wairforretail.com} \\
  \And 
  Andrea Zerio\(^*\) \\
  WAIR, Amsterdam \\%, 1059 CA \\
  \texttt{andrea@wairforretail.com} \\
  \And 
  Roy L.M. Wang \\
  WAIR, Amsterdam \\ %, 1059 CA \\
  \texttt{roy@wairforretail.com} \\
}

\begin{document}

\maketitle

\begin{abstract}

We introduce LatentTimePFN (LaT-PFN), a foundational Time Series model with a strong embedding space that enables zero-shot forecasting. To achieve this, we perform in-context learning in latent space utilizing a novel integration of the Prior-data Fitted Networks (PFN) and Joint Embedding Predictive Architecture (JEPA) frameworks. We leverage the JEPA framework to create a prediction-optimized latent representation of the underlying stochastic process that generates time series and combines it with contextual learning, using a PFN. Furthermore, we improve on preceding works by utilizing related time series as a context and introducing a normalized abstract time axis. This reduces training time and increases the versatility of the model by allowing any time granularity and forecast horizon. We show that this results in superior zero-shot predictions compared to established baselines. We also demonstrate our latent space produces informative embeddings of both individual time steps and fixed-length summaries of entire series. Finally, we observe the emergence of multi-step patch embeddings without explicit training, suggesting the model actively learns discrete tokens that encode local structures in the data, analogous to vision transformers.

\end{abstract}

%================
\section{Introduction}

Time series forecasting is a fundamental task ubiquitous to various domains, ranging from finance \citep{sezer2020financial, krollner2010financial} to healthcare \citep{kaushik2020ai, harutyunyan2019multitask}, retail \citep{fildes2022retail, wanchoo2019retail}, logistics \citep{chan2019comparison} and beyond. Traditional approaches, both statistical \cite{gardner1985exponential,contreras2003arima,taylor2018forecasting} and deep-learning-based \citep{salinas2017deepar, smyl2020hybrid}, are frequently incapable of zero-shot forecasting \cite{oreshkin2021meta} as they require training for each new dataset, limiting generalization capabilities.

In this paper, we introduce LatentTimePFN (LaT-PFN), a foundational Time Series model designed to address \emph{zero-shot forecasting} by combining the Prior-data Fitted Networks (PFN) and Joint Embedding Predictive Architecture (JEPA) frameworks. LaT-PFN is trained exclusively on a novel synthetic data generation method, allowing us to encode expert knowledge directly in the training data. Furthermore, LaT-PFN learns to perform in-context learning in latent space by approximating the Posterior Predictive Distribution (PPD). We implement this meta-learning approach by enabling the model to learn zero-shot forecasting at test time, with user-provided time series functioning as exemplary context. We train LaT-PFN using a normalized time-axis, which consolidates a wider range of time-frequency patterns, facilitating meta-learning. Additionally, this improves the model's versatility and accuracy, whilst reducing computational requirements. Following the JEPA methodology, LaT-PFN separates predicting and decoding, which are optimized independently. As such, LaT-PFN is trained to predict the next latent state and exploits a system identification loss as a regularization term, to independently improve the quality of its embedding space.

Our model demonstrates superior zero-shot prediction performance compared to baselines \citep{box2015time, dooley2023forecastpfn, taylor2018forecasting, yue2022ts2vec}, showcasing its effectiveness in handling unseen distributions. Additionally, LaT-PFN produces informative embeddings, demonstrating a comprehensive understanding of time series. Finally, we observe the emergence of multi-step patch embeddings without explicit training, suggesting that LaT-PFN actively learns discrete tokens encoding local structures in the data, reminiscent of vision transformers~\citep{dosovitskiy2020image}. Our contributions are as follows:

\begin{itemize}
    \item We introduce a novel architecture, LaT-PFN, which implements a combination of the PFN and JEPA frameworks for zero-shot predictions of univariate time series in latent space.
    \item We introduce a novel synthetic prior for simulating context-aware time series data. Specifically, our prior allows for the creation of synthetic contexts, defined as collections of time series, similar enough to enable LaT-PFN to approximate the PPD.
    \item We demonstrate, through extensive experimentation across a variety of datasets, that LaT-PFN outperforms established baselines with limited training resources, on multiple tasks.
    \item Finally, we offer an extensive analysis of the model's embedding space. This becomes particularly relevant as we notice the emergence of patch-like embeddings, without explicit training. We speculate this may be analogous to Vision Transformers (ViT) \citep{dosovitskiy2020image} and offer evidence it may even amount to a rudimentary time-series-specific corpus.
\end{itemize}

\section{Background}

\subsection{PFN: Prior-data Fitted Networks}
\label{background-PFN}

The Prior-data fitted Networks (PFN) methodology \citep{muller2021transformers}, is a meta-learning framework that explicitly trains a model for in-context learning. This is achieved through a neural network that approximates the posterior predictive distribution (PPD), thereby approximating Bayesian Inference. This requires a large volume of data, exhibiting a variety of related distributions. To facilitate this, the PFN trains on synthetic data generated by a simulation, whose parameters are sampled from a user-defined prior distribution \(Pr(\psi)\). This allows developers to explicitly encode expert knowledge on the family of datasets \(\Psi\) that the PFN is trained on, rather than relying on implicitly encoding specialized knowledge in the model's architecture \citep{muller2021transformers}. 

The PPD is typically defined as the probability distribution \(P(y_* | x_*, \mathcal{D} )\), with \(\mathcal{D} = \{(x_i, y_i)\}^N\) being a context dataset. PFNs work by approximating this distribution with network \(Q_\theta(.)\):
\begin{align*}
Q_\theta(y_* | x_*, \mathcal{D}_j) \: &\approx \: P(y_* | x_*, \mathcal{D} ) \quad
\text{with} \quad \mathcal{D}_j \sim S(\mathcal{D}_j | \psi_j), \quad \psi_j \sim Pr(\psi) 
\end{align*}

\begin{wrapfigure}{r}{0.32\textwidth}
    \vspace{-0.5cm}
    \centering
    \includegraphics[width=0.3\textwidth, height=0.3\textwidth]{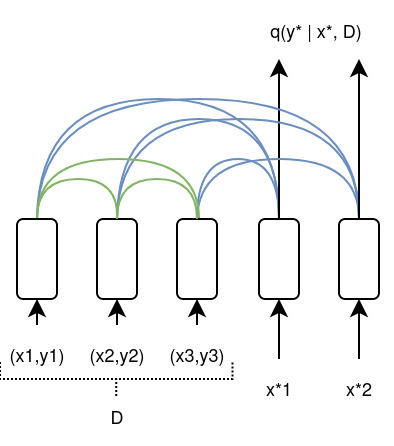}
    \caption{PFN attention \citep{muller2021transformers}}
    \label{fig:vanilla-pfn}
    \vspace{-0.5cm}
\end{wrapfigure}
Here \(Q\) is a transformer \citep{vaswani2017attention} parametrized by \(\theta\), \(S(.)\) is a simulation engine, and \(Pr(.)\) is a prior distribution over instances of the simulation parameters \(\psi \in \Psi\). Examples of the relation between features and targets are provided by the tuples \((x_i, y_i)\) within \(\mathcal{D}\), which are then passed through a transformer encoder \citep{vaswani2017attention}. The held-out features $x_*$ are passed through the decoder, with a diagonal-only (independent) target mask, allowing only attention over the encoder context (Figure~\ref{fig:vanilla-pfn}). Finally, a head and a cross-entropy loss are applied. The network \(Q\) is trained to perform meta-learning on any dataset \(\mathcal{D}_j\) \citep{muller2021transformers}. The authors prove that this is a valid approximation of the PPD -- see Appendix~\ref{sec:appendix:pfn}. Since its introduction, others have applied the PFN framework to various tasks \citep{hollmann2022tabpfn, dooley2023forecastpfn, ansari2024chronos, feuer2023scaling}. Notably, TabPFN \citep{hollmann2022tabpfn} solves small tabular datasets, zero-shot. Additionally, ForecastPFN \cite{dooley2023forecastpfn} introduced this methodology to time-series forecasting thanks to a novel synthetic prior. 

\subsection{JEPA: Joint Embedding Predictive Architecture}
\label{sec:background:JEPA}

\begin{wrapfigure}{l}{0.35\textwidth}
    \centering
    \includegraphics[width=0.28\textwidth]{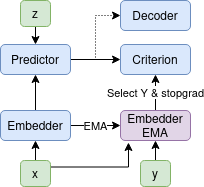}
    \caption{JEPA \citep{lecun2022path}}
    \label{fig:vanilla-jepa}
    \vspace{-0.5cm}
\end{wrapfigure}

The Joint Embedding Predictive Architecture (JEPA) \citep{lecun2022path} framework aims to create a strong representation space to model complex relationships within data, without relying on reconstruction losses or heuristics. The key concept is to predict self-supervised state transitions \(S_{t}\rightarrow S_{t+1}\) fully in \emph{latent space}, yielding embeddings that are not burdened by unnecessary detail. 
Figure~\ref{fig:vanilla-jepa} illustrates the core components, which define a sequence of transformations aimed at processing inputs within latent space. Initially, an input embedding \(\bar{x}=E_x(x)\) is produced by the input embedder. 
This is followed by a prediction \(\hat{\bar{y}}=P(\bar{x},z)\) constituting the next estimated latent state, using the transformed input \(\bar{x}\) and a variable \(z\), intended to capture relationships not observable in data space~--~for example camera angles for computer vision. A target embedding \(\bar{y}=E_y(x,y)\) is formed, taking into account both \(x\), \(y\) and their relationship. A criterion, typically a Mean Squared Error (MSE) loss \(C(\hat{\bar{y}}, \bar{y})\), evaluates the latent prediction. Finally, a generative decoder trained in isolation decodes the latent representations back to the data space \(\hat{y}=D(\hat{\bar{y}})\).

The JEPA framework has since been applied to image and video data \citep{assran2023self, bardes2023v}. The target encoder \(y_L = E_y(x, y)\) is updated as an exponential moving average (EMA) of the input encoder \(E_x\) and followed by a stop-gradient, preventing collapse by learning an identity mapping \citep{lecun2022path, assran2023self}. Other JEPA works include Motion \citep{bardes2023mc}, Audio \citep{fei2023jepa}, Point-cloud \citep{saito2024point} and EEG transfer \citep{guetschel2024s}. 

\section{Methodology}
\label{sec:method}
To the best of our knowledge, this work pioneers bringing energy-based JEPA \citep{lecun2022path} to time series forecasting, as well as combining it with the probabilistic PFN framework \citep{muller2021transformers}. Initially, this integration may appear unconventional; however, we contend that these methodologies are complementary to this modality. JEPA separates concerns of predicting and decoding. This creates an embedding space more suited to the time series modality, which is inherently stochastic. LaT-PFN can thereby explicate latent patterns representing the underlying stochastic process, prioritizing inherent predictability. However, without in-context learning, univariate time series often still lack sufficient predictive power. The context provides samples of the underlying process, enhancing predictability through contextual learning. This is particularly important for time series datasets affected by cold-start and non-stationary issues. For example, one can imagine two series with identical histories but distinct target completions; a context then provides the additional information for extrapolation.

\subsection{Problem Statement}
\label{sec:method:problem_statement}

\begin{figure}[h!]
    \centering
    \includegraphics[width=\linewidth]{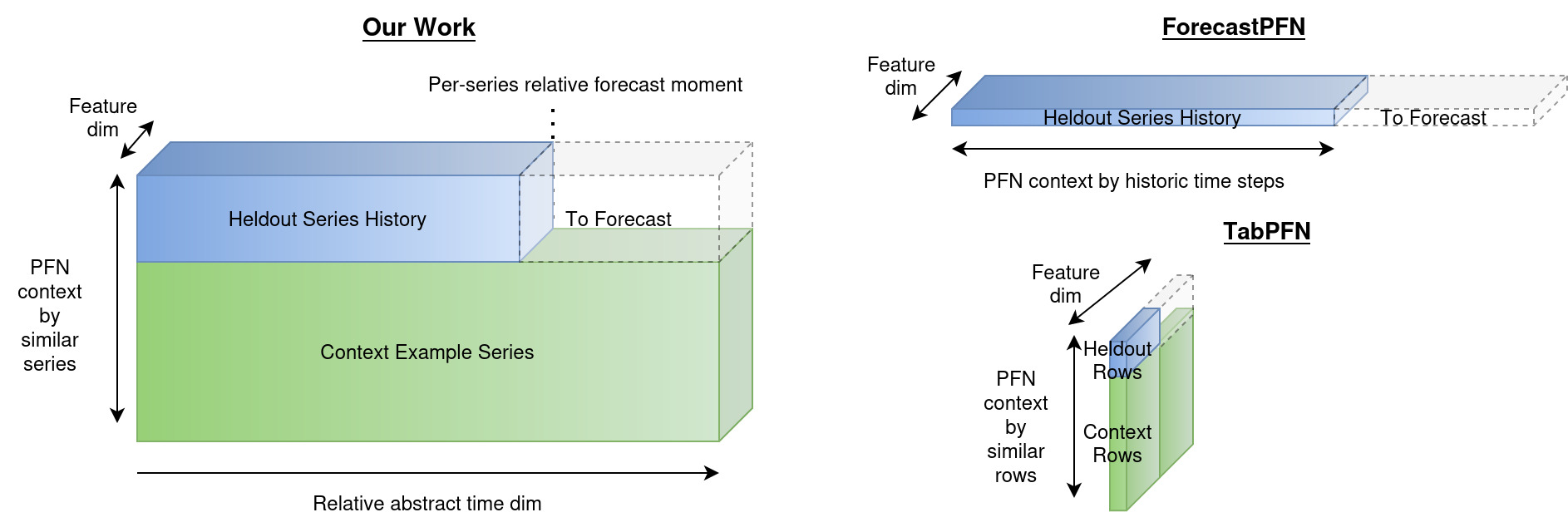}
    \caption{Our time series forecasting PFN problem statement, compared to preceding works \citep{dooley2023forecastpfn, hollmann2022tabpfn}}
    \label{fig:pfn-shapes}
\end{figure}

Starting with a common definition of time series forecasting (TSF), we seek to derive a formulation of TSF for Bayesian Inference. In this work we consider a single supervised TSF problem \(P(y|x)\) as predicting future values \(y = v_{t+1:H} = [v_{t+1}, \dots, v_{H}]\) from timestep \(t\) until horizon \(H\), conditioned on the history \( x = v_{0:t} = [v_{0}, \dots, v_{t}]\). It follows that the PPD for this problem is defined as:
\[
    \displaystyle
    P(y_* | x_*, \mathcal{D} ) \hspace{0.2em} = \hspace{0.2em} P(\:v_{*,t+1:H} \:\: | \:\: v_{*,0:t} \:, \:\: \{(x_i, y_i)\}^N ), \quad (x_i, y_i) \hspace{0.2em} = \hspace{0.2em} (v_{i,0:t}, v_{i,t+1:H} )
\]
Where \(\mathcal{D}\) is a context dataset with \(N\) examples, for \(\mathcal{D} \in \mathbb{R}^{N \times S \times F}\), sequence dimension \(S\), example dimension \(N\) and features \(F\). This differs from the ForecastPFN formulation, which defines the context \(\mathcal{D} \in \mathbb{R}^{S \times F}\) as the history of the target time series \citep{dooley2023forecastpfn}. This interpretation, although elegant, does not allow for a context for the TSF task definition and is essentially a regular TSF approach trained with simulated data. See Figure~\ref{fig:pfn-shapes} for a visualization.

To complement this notation, let embedded history \(\bar{x} = \bar{v}_{0:t}\) and latent target \(\bar{y} = \bar{v}_{t+1:H}\) be the embedded versions of history \(x=v_{0:t}\) and target \(y=v_{t+1:H}\) respectively. Similarly, let \(\hat{\bar{y}} =\hat{\bar{v}}_{t+1:H}\) be the latent forecast, and \(\hat{y} = \hat{v}_{t+1:H}\) be the forecast in data space.

\subsection{Normalized Abstract Time-axis}
Unlike \cite{dooley2023forecastpfn}, instead of using absolute time units (year, month, etc.), we map time to a fixed normalized interval \(T_0:T_H\) relative to the forecast start \(t\) of each series, with a variable sampling rate -- while ensuring the absence of any data leakage. This allows us to map context time series and past segments of the held-out series onto the same interval, independent of their actual place in history. The approach aligns well with the idea of using simulated data to learn Bayesian inference, as it creates a universal and consistent time dimension in which the model can more readily learn general temporal patterns. Furthermore, reducing the number of temporal degrees of freedom that require training, leads to a more efficient convergence and decreases the computational demands for training.

\subsection{An Architecture for Latent In-context Forecasting}
\begin{figure}[h]
    \centering
    \includegraphics[width=\textwidth]{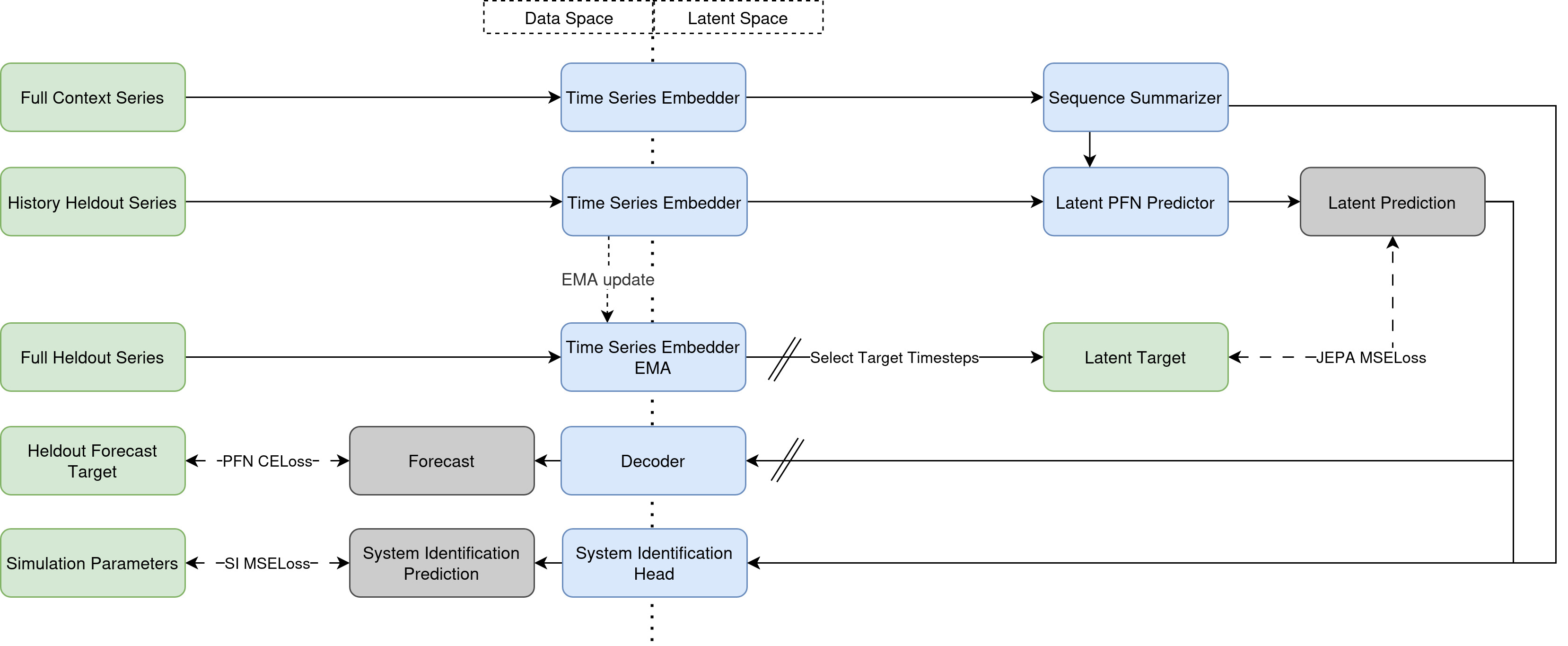}
    \caption{The LaT-PFN Architecture. The context is embedded in fixed-length series-vectors. These are fed into the PFN Predictor transformer, with the embedded held-out history prompts, using cross-attention. The latent predictions are compared to the latent target, then decoded with a stop-gradient, and compared to real targets. Finally, we apply a supervised regularization on simulation parameters.}
    \label{fig:architecture}
\end{figure}
\paragraph{Embedder}
\begin{wrapfigure}{l}{0.22\textwidth}
    \vspace{-0.5cm}
    \includegraphics[width=0.2\textwidth]{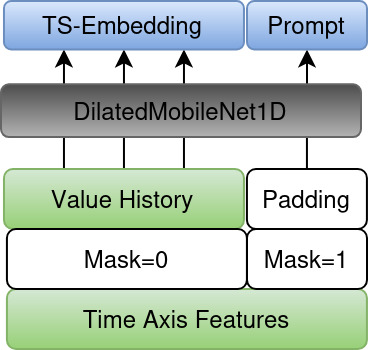}
     \caption{Production of embeddings and prompts by masking and selection}
    \label{fig:convtention}
    \vspace{-0.75cm}
\end{wrapfigure}

The embedder \(E_\theta\) is an eight-layer dilated Mobilenet1D \citep{howard2017mobilenets}. We empirically found convolutions to be superior to attention over the sequence dimension, in line with \citep{yue2022ts2vec, liu2023itransformer}. The component is responsible for feature extraction along the sequence dimension. First, we create the context embeddings \(\bar{\mathcal{D}}\) and the latent target \(\bar{y}\) with time \(T\) and value \(V\) features. For the latent target specifically, we use an EMA of the embedder parameters, apply a stop-gradient, and select only the future timesteps:
\begin{align*}
    \displaystyle
    \\[-12pt]
    \bar{\mathcal{D}}  \hspace{0.5em} &=  \hspace{0.5em} E_\theta(\mathcal{D})  \\[4pt]
    \bar{y}  \hspace{0.5em} &=  \hspace{0.5em} Select_{t+1:H}\big[\:StopGrad\big(E_{\theta_\texttt{EMA}}(x, y)\:\big)\:\:\big]
\end{align*}
\\Next, we create the held-out embeddings \(\bar{x}\) and the prompt \(\bar{z}\):
\[
    \bar{x}, \bar{z}  \hspace{0.5em} =  \hspace{0.5em} E_\theta(x)
\]
 To avoid introducing forward-looking values, we apply padding and masking to the input features. The prompt is therefore a function of the historical values and the entire time axis (including the horizon). See Figure~\ref{fig:convtention} for an illustration of the aforementioned process. 

\paragraph{Predictor}
\label{sec:method:predictor}

The predictor \(\textit{PFN}_\theta\) forecasts the next latent state, given the prompt and context. We apply learned average pooling \(\textit{AVG}_\theta\) over the sequence dimension of the embedded context \(\bar{\mathcal{D}}\), by cross-attention with a learned vector \(\vec{q} \in \theta\). These are then provided to the encoder of the latent PFN predictor in Figure~\ref{fig:architecture}, with self-attention over the examples dimension. 
\[
    \bar{\eta} \hspace{0.5em} =  \hspace{0.5em} \textit{AVG}_\theta(\bar{\mathcal{D}}) 
\]
In order to apply 2D attention over the held-out series, we flatten the examples dimension \(N\) and sequence dimension \(S\) of the prompt \(\bar{z}\). Next, we pass it to the decoder, whilst ensuring adjacent sequences from held-out series are not dependent on sharing information amongst each other. 
\[
    M \hspace{0.33em} = \hspace{0.33em} diag(1)^{N \cdot S \times N \cdot S}, \quad
    \hat{\bar{y}} \hspace{0.33em} =  \hspace{0.33em} \textit{PFN}_\theta(\bar{z}, \bar{\eta}, M), \quad  \mathcal{L}_\text{latent} \hspace{0.33em} = \hspace{0.33em} \frac{1}{n} \sum^n (\hat{\bar{y}} - \bar{y})^2
\]
This is achieved by the diagonal-only mask \(M\), which we adopt from the original PFN formulation \citep{muller2021transformers}. Consequently, we stay true to the design from Figure~\ref{fig:vanilla-pfn}.

\paragraph{Decoder}

The decoder \(D_\theta\) is a three-layer feedforward network tasked with decoding latent predictions, approximating the PPD in data space (see Appendix~\ref{sec:appendix:pfn}). It is trained with independent gradients in alignment with the JEPA methodology \citep{lecun2022path, assran2023self, bardes2023v}. We empirically found this to be critical for generating stable embeddings upstream. The decoder utilizes a cross-entropy loss to define a categorical distribution over 100 segmented bins of the output space (see Appendix~\ref{sec:appendix:pfn}), incorporating label smoothing of 0.01 \citep{szegedy2016rethinking}. This is made feasible by z-normalizing the values \(V\) with 2-std, towards a fixed interval. Consequently, the decoder yields an output without making any assumptions on the target distribution -- adhering closely to the PFN formulation \citep{muller2021transformers}. 
\[
    \displaystyle
    \hat{y} \hspace{0.33em} = \hspace{0.33em} D_\theta \big (\:StopGrad(\hat{\bar{y}}) \:\big), \quad
    \mathcal{L}_\text{decoder} \hspace{0.33em} = \hspace{0.33em} \mathbb{E}_{p(y)} \log p(\hat{y} \: | \:\: \hat{\bar{y}})
\]

\paragraph{System Identification Head}

Inspired by Sim2Real research (e.g. \citep{yu2019sim}), we add a regularization term, to encourage consistency between history and forecast. This component processes the summarised vectors of both context and predicted held-out series, to identify the simulation parameters that generated the context \(\mathcal{D}\). This approach further supervises the latent space, directing it to the underlying characteristics of the simulation, such as trend and seasonality. This, in turn, leads to improved forecasting capabilities and should stabilize training \citep{lecun2022path}. The latent embeddings and predictions are concatenated and averaged, then passed to a head \(H_\theta\) alongside the embedded context. \(H_\theta\) learns a multi-target regression towards unit-normalized simulation parameters.
\[
    \hat{y}_\text{si} \hspace{0.33em} = \hspace{0.33em} H_\theta \big ( \: \textit{AVG}_\theta(\hat{\bar{y}}, \bar{x}),  \bar{C} \: \big ),  \quad
    \mathcal{L}_\text{si} \hspace{0.33em} = \hspace{0.33em} \frac{1}{n} \sum^n (\hat{y}_\text{si} - \psi_j)^2
\]

\paragraph{Loss}
We define our loss as follows:
\[
\mathcal{L} \hspace{0.2em} = \hspace{0.2em} \lambda_\text{latent} \mathcal{L}_\text{latent} + \lambda_{si} \mathcal{L}_\text{si} + \mathcal{L}_\text{decoder}, \quad \lambda_\text{latent}=3.77e^{-3}, \quad \lambda_{si}=1e^{-7}
\]
The decoder is trained following a stop-gradient operation, ensuring one-way independence.

\subsection{Context-aware Synthetic Prior with Triple Sampling}
\label{sec:methodology:simulation}
\begin{figure}[]
  \centering
  \captionsetup[subfigure]{justification=centering, labelformat=empty}

\subfloat[\textit{\footnotesize ECL}]{
  \includegraphics[trim={5cm 27cm 27cm 5cm},clip,width=0.28\linewidth]{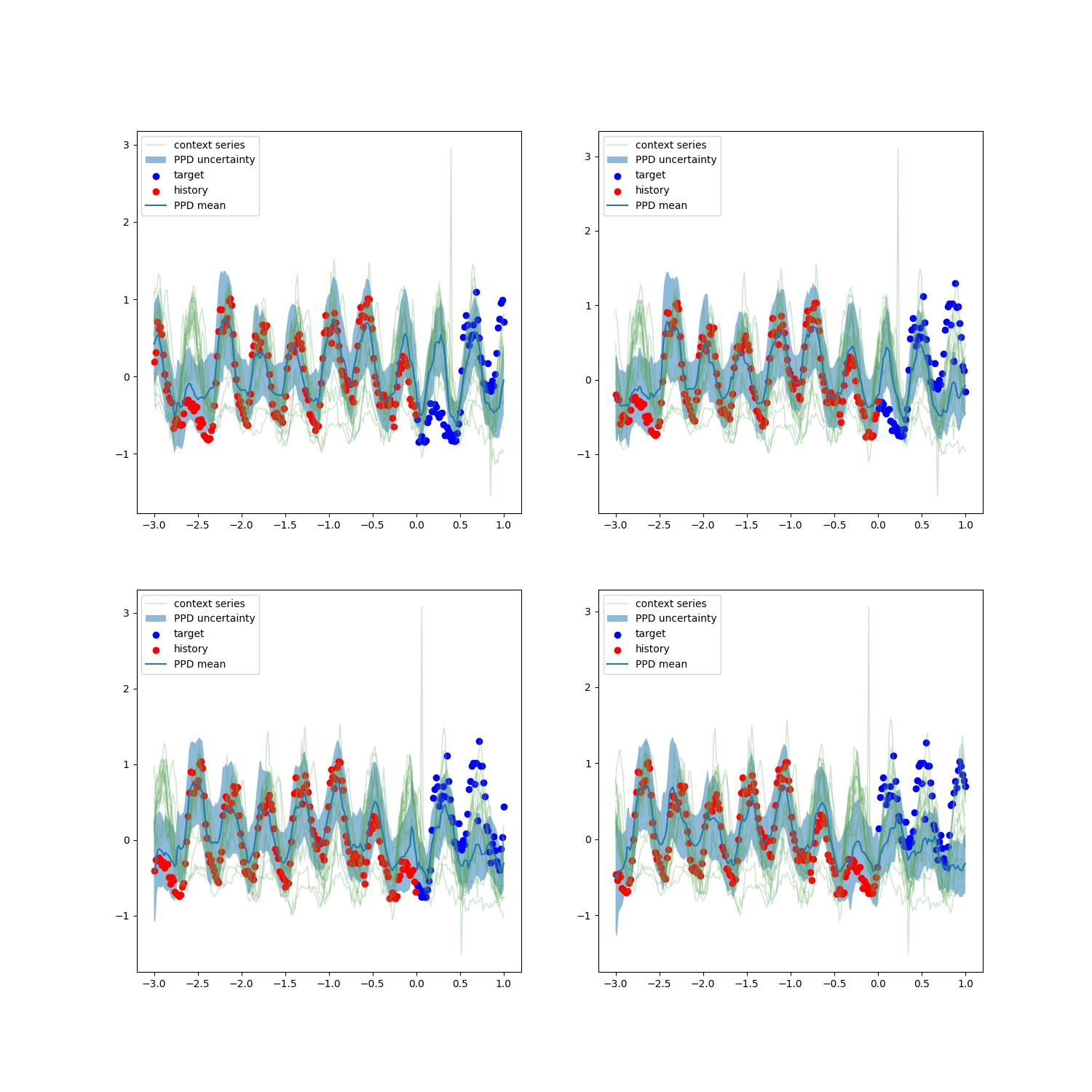} \label{fig:fits:todo}
}
\hfill
\subfloat[\textit{\footnotesize Traffic }]{
  \includegraphics[trim={5cm 27cm 27cm 5cm},clip,width=0.28\linewidth]{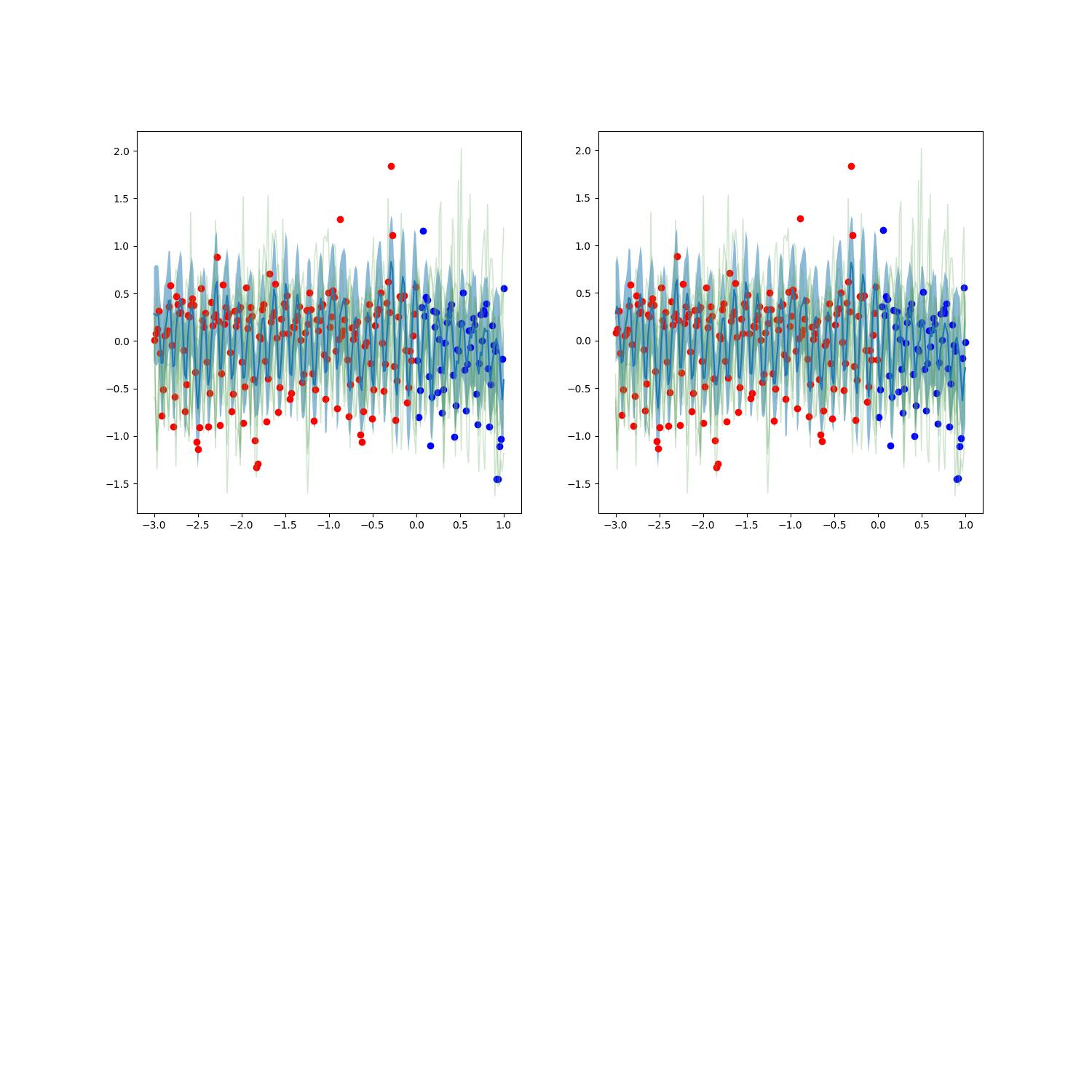} \label{fig:fits:traffic}
}
\hfill
\subfloat[\textit{\footnotesize Illness}]{
  \includegraphics[trim={27cm 5cm 5cm 27cm},clip,width=0.28\linewidth]{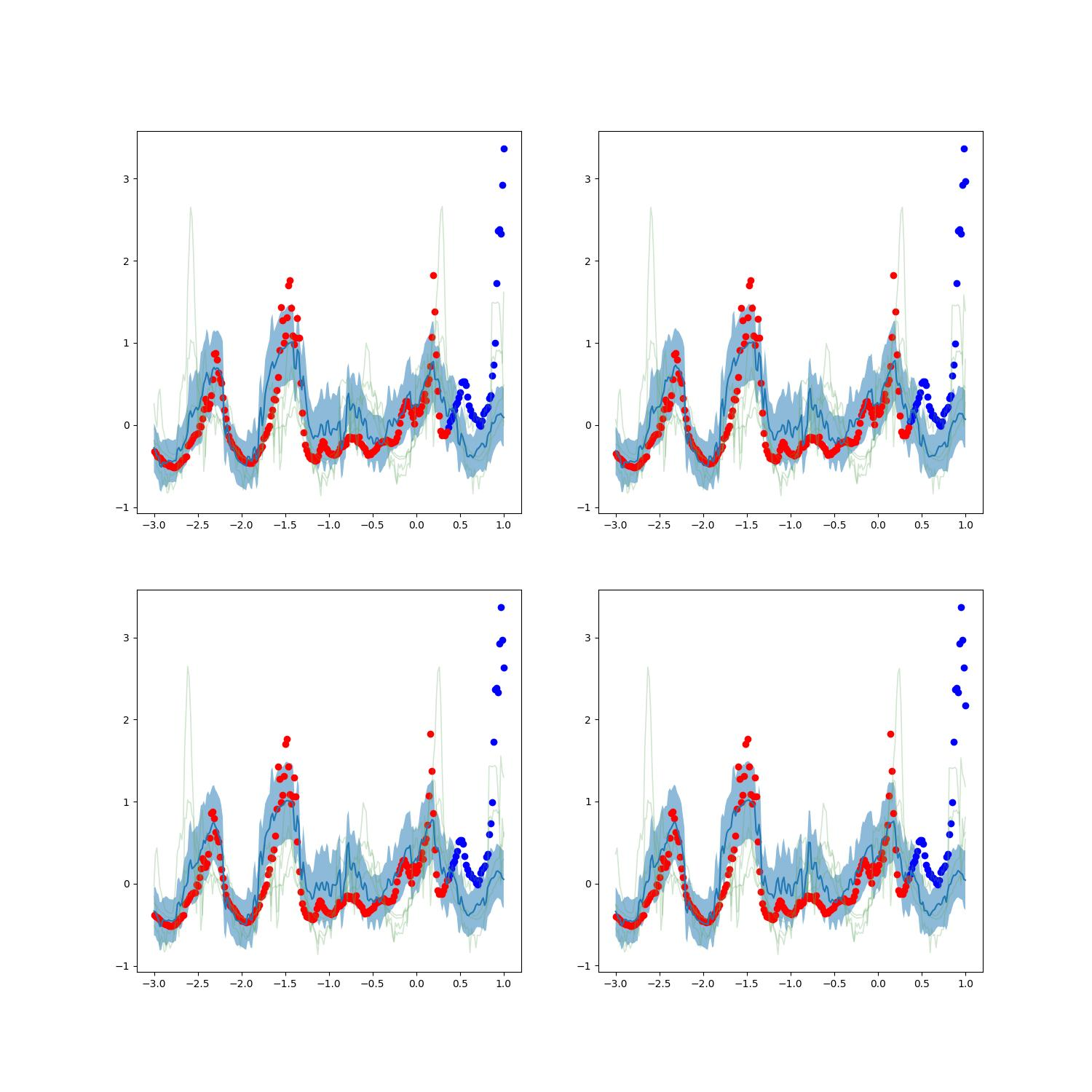} \label{fig:fits:illness}
} \vspace{-0.2cm}\\
\subfloat[\textit{\footnotesize Etth1 }]{
  \includegraphics[trim={27cm 27cm 5cm 5cm},clip,width=0.28\linewidth]{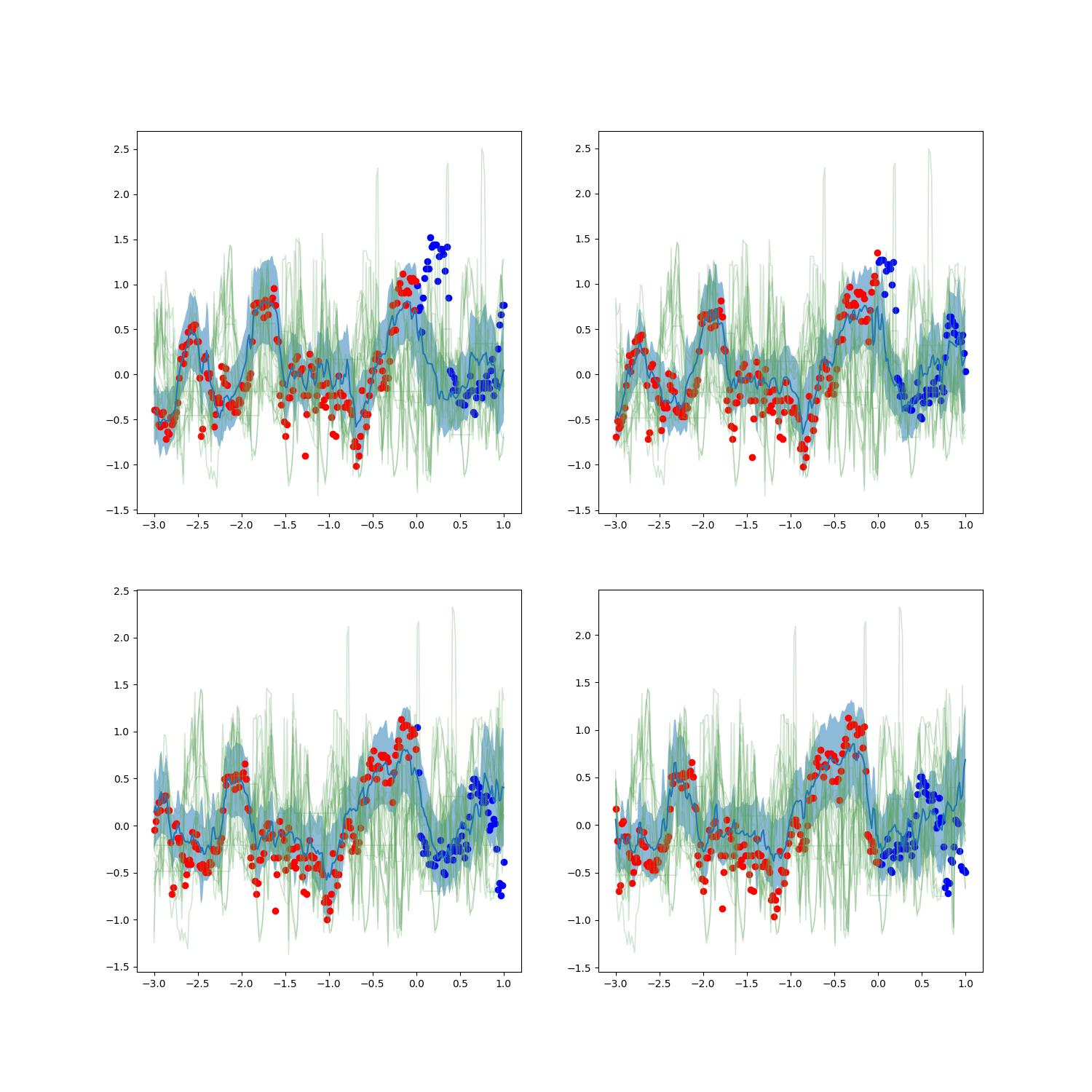} \label{fig:fits:etth1}
}
\hfill
\subfloat[\textit{\footnotesize Etth2 }]{
  \includegraphics[trim={27cm 5cm 5cm 27cm},clip,width=0.28\linewidth]{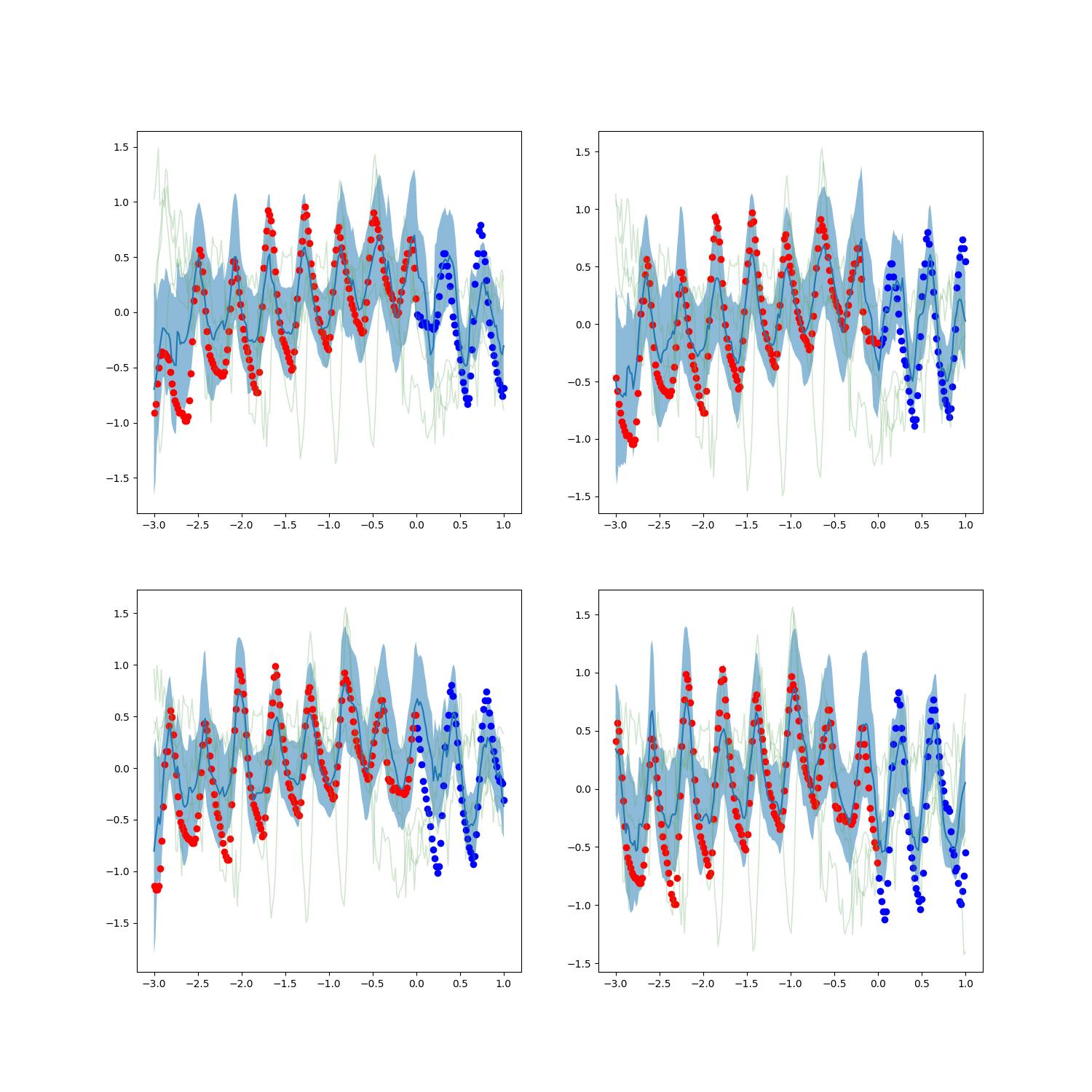} \label{fig:fits:etth2}
}
\hfill
\subfloat[\textit{\footnotesize Synthetic Prior}]{
  \includegraphics[trim={5cm 5cm 27cm 27cm},clip,width=0.28\linewidth]{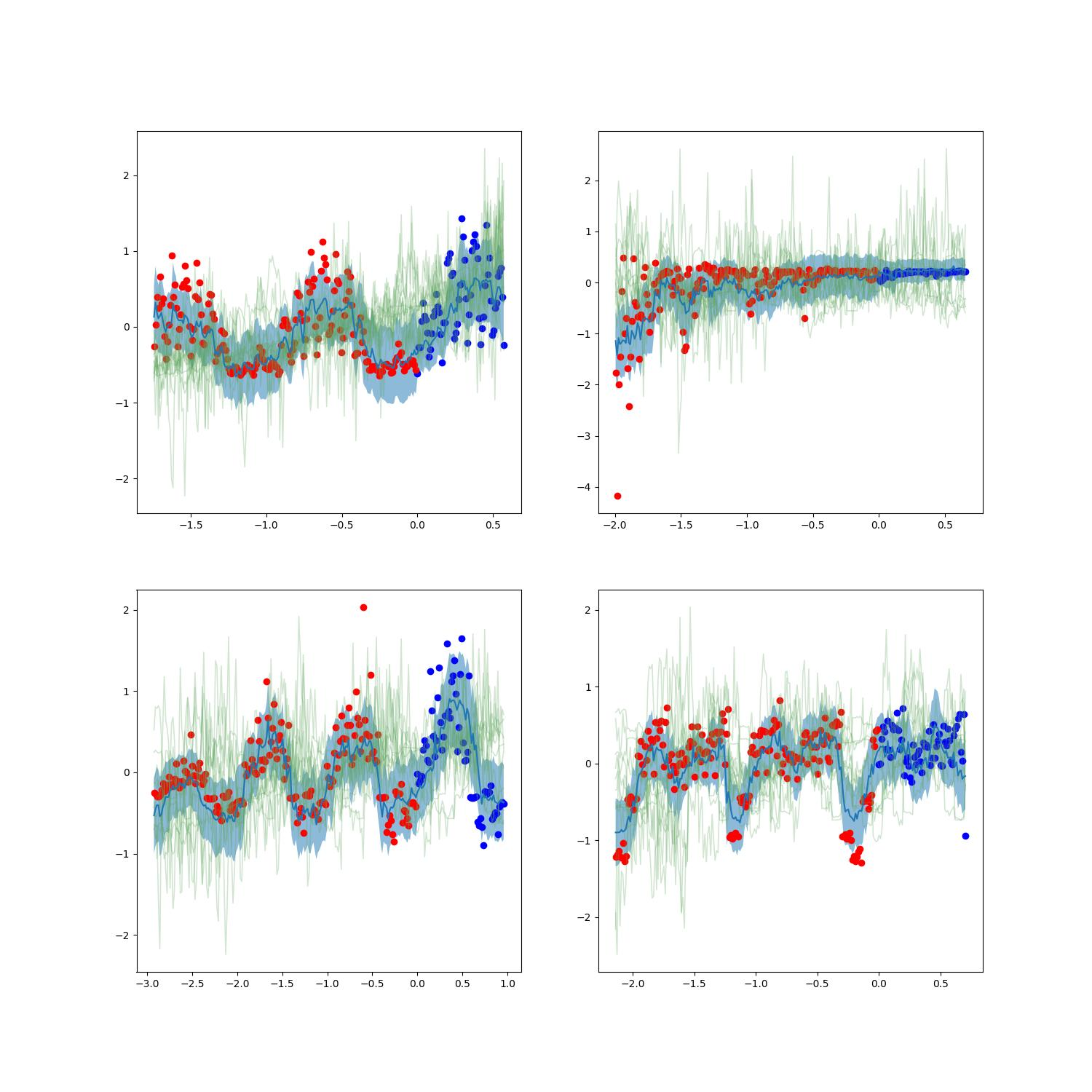} \label{fig:fits:sim}
}
  \caption{Decoded fits plotted on the target, with history and target in blue and red scatter respectively, as well as the PPD fore- and backcast mean as a blue line with variance, and finally the input example context in green in the background. The fits demonstrate the predictions in latent space have picked up concepts such as seasonality and trend, making it easier to extrapolate after the forecast moment.}
  \label{fig:fits}
\end{figure}

Following the PFN example \citep{muller2021transformers, dooley2023forecastpfn, hollmann2022tabpfn}, Lat-PFN was trained exclusively on synthetic data. For this work, we adopted and adjusted ForecastPFN's time series synthetic prior \citep{dooley2023forecastpfn}. Like ForecastPFN, we set hyperprior parameters that govern the sampling of simulation parameters. Unlike ForecastPFN however, our context is defined not by the history of a single time series, but by a collection of example time series. Hence, for most simulation parameters, we introduce a context dimension by employing a three-step sampling strategy:
\begin{align*}
\displaystyle
    \alpha_c \hspace{0.3em} \sim \hspace{0.3em} &\mathcal{U}(\alpha^*) \\
    \mu_1, \mu_2 \hspace{0.3em} \sim \hspace{0.3em} &\mathcal{U}(\alpha_c), \\
    \psi_i \hspace{0.1em} \sim \hspace{0.2em} \mathcal{N}(\mu_1, \sigma^*), \quad &\psi_j \hspace{0.1em} \sim \hspace{0.2em} \mathcal{N}(\mu_2, \sigma^*), \quad i,j \in [0,N]
\end{align*}
Specifically, for any simulation parameter in $\psi_i$ the procedure is defined as:
\begin{enumerate}
    \item \textbf{Hyperprior Parameters} $(\alpha^*)$: Set the global limits for context synthesis.
    \item \textbf{Context Parameters} $(\alpha_c)$: Ranges sampled uniformly from the hyperprior to define the limits of an individual context.
    \item \textbf{Sub-context Parameters} $(\mu)$: Cluster centers $(\mu_1, \mu_2)$, sampled uniformly from the context range, are used to introduce contrast within a context and guide the model to recognize when a context example may be non-informative.
    \item \textbf{Series Parameters} $(\psi)$: Point estimates normally sampled with fixed hyperprior variance~\(\sigma^*\), used as simulation parameters in $S(v_{i,0:T} | \psi_i)$ to generate individual training series.
\end{enumerate}

This approach ensures balance in inter- and intra-context variance, promoting coherence without trivializing the series and enhancing the model's ability to generalize in a zero-shot setting. A full description of the synthetic prior and which parameters are used can be found in Appendix~\ref{sec:appendix:reproducibility:simulation}.

\section{Experimental Setup and Training Details}
\label{sec:experimental-details}

\paragraph{Training}  During training, we employ a linear warmup schedule for weight decay, and target encoder ema decay -- in line with previous JEPA work \citep{assran2023self, bardes2023v}. We trained the final model for 24 hours on a single NVIDIA A10G Tensor Core GPU -- per seed. Additionally, we employ tuning on schedule parameters and \(\mu\)-parametrization, which was essential for training stability \citep{yang2022tensor}. For more training and tuning details, see Appendix~\ref{sec:appendix:reproducibility}.

\paragraph{Metrics}  We report the mean and error statistics, assumed normally distributed, of the zero-shot performance of five seeds. These are accuracy, MSE per time-step, and the cumulative relative root mean squared error (CRRMSE). The latter is normalized by the cumulative target. We motivate CRRMSE for its scale-invariance and practical applicability in tasks such as sales forecasting. 

\paragraph{Benchmarks \& Context Curation} For real datasets, contexts were divided in fixed windows, normalized to a fixed time interval, and Z-normalized with 2-std using the history. The context-sample selection process is coined \emph{Context Curation}, and is analogous to Prompt Engineering in Large Language Models (LLMs) \cite{marvin2023prompt, white2023prompt}. For dataset-specific details see Appendix~\ref{sec:appendix:reproducibility}.

\paragraph{Baselines} We selected FBProphet \citep{taylor2018forecasting}, ARIMA \citep{box2015time}, and ForecastPFN \citep{dooley2023forecastpfn} as forecasting baselines, all of which have open-source codebases. We compared against TS2Vec, in a downstream classification task, \citep{yue2022ts2vec} to evaluate the quality of the embedding space. We use the original authors' codebase and either download the trained model weights or re-train, wherever applicable. Finally, all baselines are tuned, where applicable, and then evaluated on identical datasets. For more details on baselines, see Appendix~\ref{sec:appendix:reproducibility}.

\begin{table}[]
\centering
\caption{Forecasting scores (with std) for different time horizons. Results demonstrate a strong performance for all of the benchmarks from decoded latent predictions. See data details in Appendix~\ref{sec:appendix:reproducibility}}
\label{tab:forecast}
\begin{tabular}{@{}llllll@{}}
\toprule
\multicolumn{6}{c}{\textit{Mean Squared Error (MSE)}} \\
                & ECL                         & ETTh1                     & ETTh2                     & Illness                     & Traffic                     \\ \midrule
                                                                      
Arima   & 0.44                                 & \textbf{0.44}                                 & \textbf{0.45}                                 & 0.50                                 & 0.25                                 \\
ForecastPFN            & 1.16                                 & 0.84                                 & 0.74                                 & 3.05                                 & 0.24                                 \\
FBProphet & 0.61                                 & 0.98                                 & 0.67                                 & 0.96                                 & 0.29                                 \\
LaT-PFN         & \textbf{0.32} { \scriptsize \(\pm\) 4.1e-2 } & 0.59 { \scriptsize \(\pm\) 3.0e-2 } & \textbf{0.45} { \scriptsize \(\pm\) 1.8e-2 } & \textbf{0.23} { \scriptsize \(\pm\) 2.3e-2 } & \textbf{0.23} { \scriptsize \(\pm\) 2.8e-2 } \\ \midrule
\multicolumn{6}{c}{\textit{Cumulative Relative Root Mean Squared Error (CRRMSE)}}                                                                                                                                                                                         \\ \midrule
Arima   & 5.83                                 & 4.93                                 & 4.74                                 & 5.02                                & 2.68                                 \\
ForecastPFN            & 11.39                                & 7.07                                 & 6.66                                 & 15.6                                & 4                                    \\
FBProphet & 5.75                                 & 8.33                                 & 6.42                                 & 7.79                                 & 3.34                                 \\
LaT-PFN         & \textbf{4.03} { \scriptsize \(\pm\) 2.6e-1 } & \textbf{4.75} { \scriptsize \(\pm\) 2.1e-1 } & \textbf{3.67} { \scriptsize \(\pm\) 7.2e-1 } & \textbf{3.11} { \scriptsize \(\pm\) 1.8e-1 } & \textbf{2.01} { \scriptsize \(\pm\) 1.9e-1 } \\ \bottomrule
\end{tabular}
\end{table}

\section{Results \& Discussion}
\label{sec:results}

\paragraph{Forecasting}

\begin{wraptable}{r}{0.45\linewidth}
\vspace{-0.35cm}
\centering
\caption{Overall accuracy (with std) of all UCR datasets, by fitting SVM on top of fixed-length summary embeddings \citep{UCRArchive2018}. Per-dataset scores are in Appendix~\ref{sec:appendix:results}.}
\label{tab:classification}
\begin{tabular}{@{}ccc@{}}
\toprule
                             & train budget & accuracy        \\ \midrule
TS2Vec                       & 35 epochs & 44.8\%{ \scriptsize \(\pm\) 1.1\% }        \\
\multicolumn{1}{c}{LaT-PFN} & \textbf{zero-shot}       & \textbf{47.9}\%{ \scriptsize \(\pm\) 3.7\% }\\ \bottomrule
\end{tabular}
\end{wraptable}

We find that LaT-PFN exhibits strong, zero-shot forecasting performance across a wide variety of datasets, containing very different temporal patterns and levels of noise. We outline the results of the zero-shot forecasting experiments in Table~\ref{tab:forecast}. As can be observed, in nearly all cases LaT-PFN outperforms the baselines. Furthermore, we demonstrate visual curve-fits in Figure~\ref{fig:fits}. Here we can see the model understands useful patterns from the context, such as trend and seasonality, and uses them to reach a good extrapolation for the held-out targets. This suggests that LaT-PFN is inherently strong at generalizing across distributions for zero-shot forecasting, which makes it widely applicable for downstream tasks.
\begin{wrapfigure}{l}{0.24\textwidth}
    \centering
    \vspace{-0.2cm}
    \includegraphics[width=0.24\textwidth, height=0.24\textwidth]{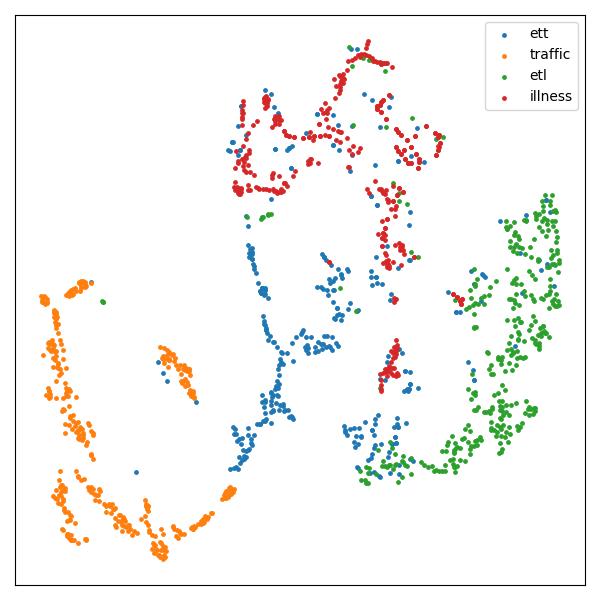}
    \caption{T-SNE \citep{van2008visualizing} of fixed-length embeddings by dataset}
    \label{fig:tsne}
    \vspace{-16pt}
\end{wrapfigure}

Next, we investigate the importance of the context by evaluating the uplift in performance on a synthetic validation set, for different context sizes. We chose to carry out this experiment on simulated data to guarantee a controlled environment. The result is visualized in Figure~\ref{fig:ablation}, which shows a clear correlation between performance and context size. More interestingly, this pattern persists even beyond the original context size of the model during training (represented by the vertical dotted line). This validates our assumption that the key to carrying out in-context learning for zero-shot forecasting is to provide a rich and substantial set of examples. This can be especially impactful when applied in a production setting with an abundance of example series, such as sales forecasting.
\begin{figure}[h]
    \centering
    \includegraphics[width=0.8\textwidth]{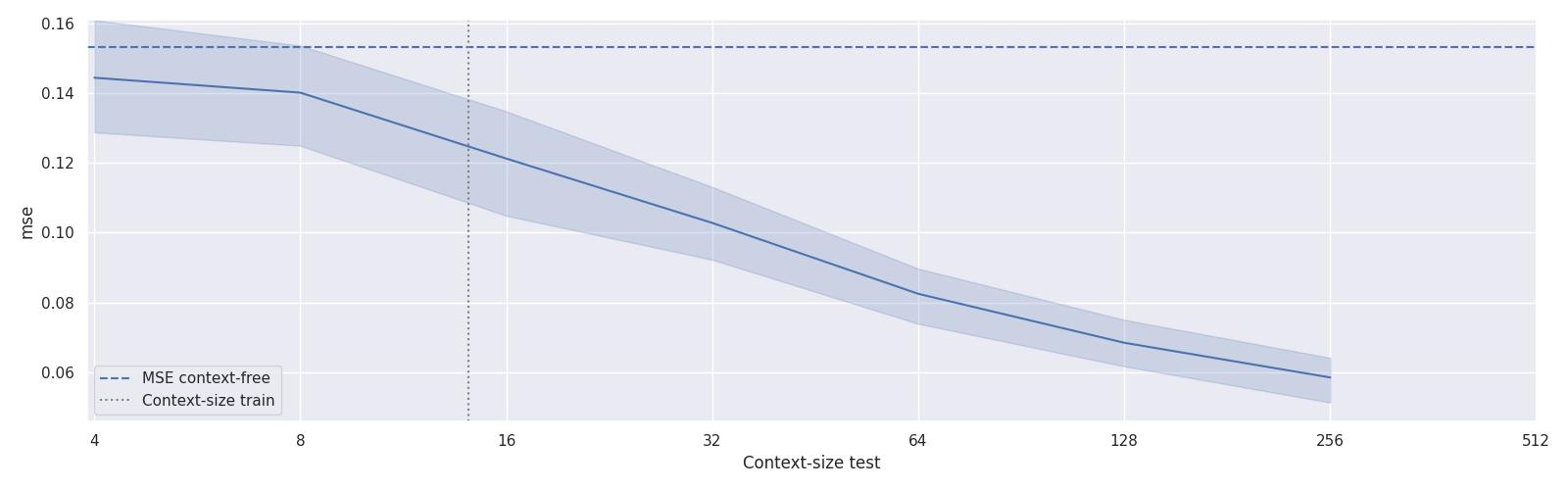}
    \caption{Extrapolating evaluation context generalizes beyond training context size}
    \label{fig:ablation}
\end{figure}
\paragraph{Classification} 

We evaluate the quality of LaT-PFN's embedding space by introducing a second downstream task, namely time series classification on the 128 multi-domain UCR datasets \citep{UCRArchive2018}. We compare our model to TS2Vec, a state-of-the-art universal representation learning framework for time series \citep{yue2022ts2vec}. We train TS2Vec on the UCR datasets, then fit Support Vector Machines (SVM) \citep{bishop2006pattern} on their embedding space. We compare this to an SVM fitted on the fixed-length summary embeddings of a frozen LaT-PFN, trained on synthetic data only. As is evident in the results reported in Table~\ref{tab:classification}, LaT-PFN outperforms TS2Vec on the combined datasets, despite the former being run in the zero-shot setting. See Appendix~\ref{sec:appendix:results} for individual dataset scores. 

We argue these results may be due to bias introduced by time series-specific heuristics and contrastive positive pair selection, both of which TS2Vec and similar methods rely heavily upon \citep{yue2022ts2vec, yang2022timeclr}. By contrast, LaT-PFN independently discovers a representative embedding space in which self-supervised learning is highly efficient. Furthermore, TS2Vec is trained on the datasets it is evaluated on. This may traditionally be thought of as an advantage. Yet, often these datasets are small and lack variety, increasing the risk of out-of-distribution data in the test set. On the other hand, LatPFN's prior ensures a wider variety of contexts, whilst adapting to the specific context during test time.

\paragraph{Exploring the Embedding Space}
\label{sec:results:explore}
\begin{wrapfigure}{r}{0.36\textwidth}
    \addtocounter{figure}{1}
    \vspace{-0.4cm}
    \centering
    \includegraphics[width=0.35\textwidth, height=0.25\textwidth]{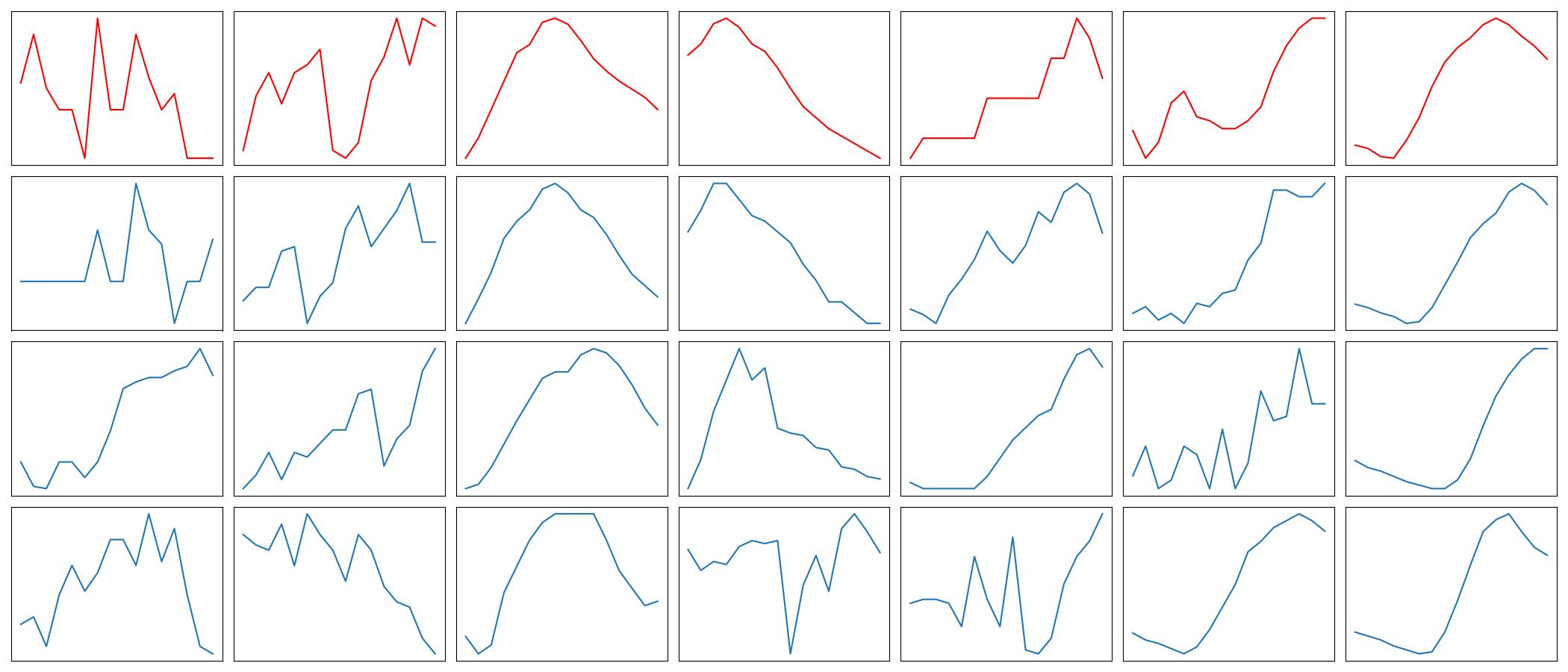}
    \caption{Found patches (red) and top-3 closest matches (blue)}
    \label{fig:path_neighbours}
    % \vspace{-1cm}
\end{wrapfigure}

We carry out a qualitative analysis of the latent space generated by LaT-PFN. When analyzing the fixed-length summary embeddings after dimensionality reduction, we observe a clear separation in clusters by dataset -- see Figure~\ref{fig:tsne}. This suggests that LaT-PFN picks up distributional differences between these domains, making the fixed-length summary embeddings useful for search, classification, and other downstream applications.

\begin{figure}
    \addtocounter{figure}{-2}
    \centering
    \includegraphics[width=\linewidth]{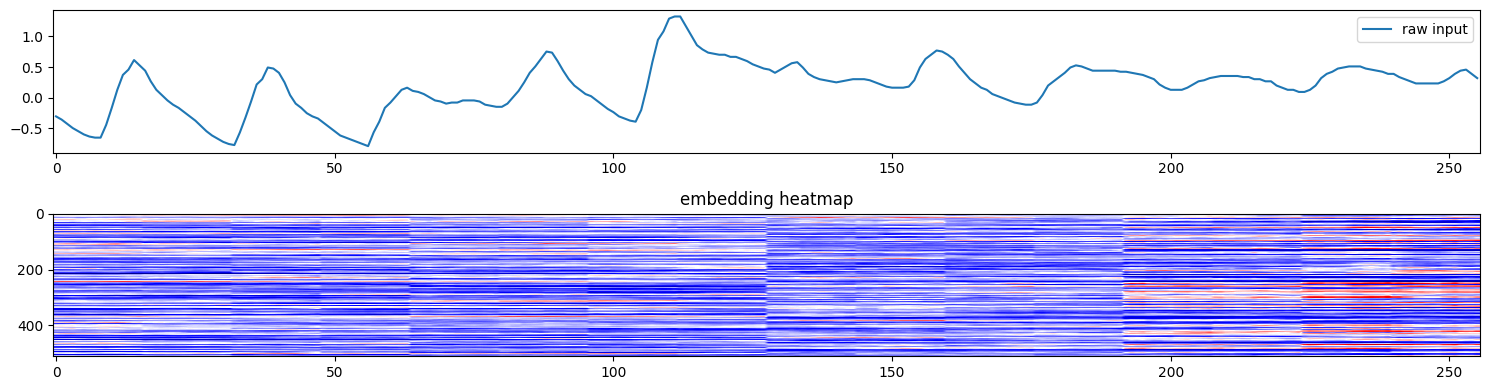}
    \caption{Normalized embeddings across sequence show patch-like behavior. More in Appendix~\ref{sec:appendix:latent_space}}
    \label{fig:embedding-ts}
\end{figure}

Interestingly, in the per-timestep embeddings we can observe the emergence of distinct, regular, and discrete patch-like tokens in the heatmaps of the embedding space. An example of this is visualized in Figure~\ref{fig:embedding-ts}. Whilst not always humanly interpretable, discernible patterns often emerge from a closer analysis of these patches. For instance, Figure \ref{fig:path_neighbours} plots some examples in the data space, alongside their closest matches when taking the \(L_2\)-distance between the average patch-embedding. We can observe that these are visually related and seem to describe specific local features.  This presents similarities to the patch-like processing of visual "tokens" in Vision Transformers (ViT) and their approach to high-level feature representation \cite{dosovitskiy2020image}. LaT-PFN, however, appears to have learned these tokens independently, without the need to outright encode this pattern in the model's architecture.

Finally, after applying dimensionality reduction to the set of patch embeddings (Figure~\ref{fig:tsne_path_clusters}), we notice the clear emergence of a few clusters. Considering the noted similarity between the shapes of patches in data space, and their corresponding averaged patch "token" embeddings, we hypothesize that these patches may represent a latent \emph{vocabulary} independently learned by the model, comprising common "words" across multi-domain series. This would once more validate our core assumption that, despite time series data often being perceived as highly domain-dependant, there is in fact enough commonality to warrant the use of meta-learning approaches to create foundational models for this data modality. 

\section{Related Work}
\begin{wrapfigure}{1,r}{0.28\textwidth}
    \addtocounter{figure}{1}
    \centering
    \vspace{-0.5cm}
    \includegraphics[width=0.2\textwidth, height=0.2\textwidth]{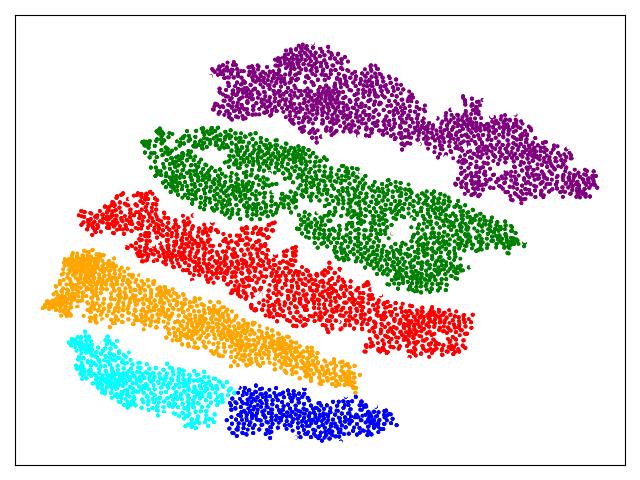}
    \caption{DBScan \citep{ester1996density} clustering of patches.}
    \label{fig:tsne_path_clusters}
    \vspace{-0.5cm}
\end{wrapfigure}

For time series forecasting, the focus has recently shifted from statistical models such as ARIMA \citep{box2015time}, ES \citep{gardner1985exponential} and FBProphet \citep{taylor2018forecasting} towards deep learning. Early pioneers are DeepAR \citep{salinas2017deepar} and hybrid methods \citep{smyl2020hybrid}. Subsequently, N-Beats \citep{oreshkin2019n}, N-hits \citep{challu2023nhits}, Informer \citep{zhou2021informer} and FEDformer \citep{zhou2022fedformer}, combine classical with deep learning paradigms.

Following the success of the NLP domain, many attempts have been made to leverage the transformer architecture \citep{vaswani2017attention} to create foundation models \citep{bommasani2021opportunities} for time series forecasting. The two main approaches for doing so can be categorized as; (a) re-purposing LLMs \citep{gruver2024large, zhou2024one, jin2023time} and (b) training transformer models from scratch \citep{dooley2023forecastpfn, ansari2024chronos, woo2024unified, das2023decoder,yeh2023toward, rasul2023lag}. By contrast, our research seeks to integrate representation- and in-context learning in time series forecasting, combining strong performance with a low computational and data budget.

Another approach altogether is to create embeddings and leave the downstream application to the discretion of end users. Notable are TS2Vec \citep{yue2022ts2vec}, TimeCLR \citep{yang2022timeclr}, SimMTM \citep{dong2024simmtm} and  TF-C \citep{zhang2022tfc}. These methods often rely heavily on expert heuristics, whereas our approach serves both as an embedder and a forecaster, leveraging a self-supervised latent space to drive predictions.

\section{Limitations}
\label{sec:limitations}
The research presented has a few limitations. Firstly, the current framework is constrained to univariate series, limiting its applicability to multivariate scenarios. Similarly, this model has yet to prove effective in handling hierarchical, discrete, or zero-inflated series, which are often underrepresented in this research field. Defining appropriate synthetic priors could address this, which we leave to future work. Secondly, we observe a low level of standardization for time series deep learning research. Whilst we found an (over) abundance of standard datasets, we observed that data processing, and target and horizon selection are still arbitrary, reducing the comparability of results across studies. This is beyond the scope of this paper but may be addressed in future work. Thirdly, we designed our model to be highly adaptive to a given context. As a result, sub-optimal contexts can lead to variability in performance. We argue this is analogous to prompt engineering, which assigns this responsibility to the expert user. In future work, we hope to address this by exploring automated methods -- via retrieval augmented generation (RAG) \citep{lewis2020retrieval} or prompt tuning \citep{lester2021power}. Finally, the model has proven somewhat sensitive to initialization and the normalization functions used. Addressing these points will be valuable for enhancing the model's versatility.

\section{Conclusion}
In this paper, we proposed LaT-PFN, a novel architecture for zero-shot univariate time series forecasting. The model integrates both the PFN and JEPA architecture to allow for in-context learning in latent space. LaT-PFN is trained exclusively on synthetic data, for which we define a novel prior. Combined with a normalized abstract time-axis, this makes our model extremely versatile and able to achieve strong forecasting results across different data distributions, time granularities, and forecast horizons. Finally, its strong embedding space exhibits interesting emergent qualities, such as the development of patch-like tokens that resemble a corpus of time series features. This makes LaT-PFN ideal for supporting transfer learning on downstream models.

\FloatBarrier

\newpage

\begin{ack}
This research was fully funded by WAIR (AI4R B.V.). We would like to extend our sincere gratitude to our team and all contributors whose invaluable assistance and insightful suggestions made this work possible.
\end{ack}

{
    \small
    \bibliographystyle{abbrvnat} 
    \bibliography{references}
}

\newpage

\appendix

\section*{Appendix / supplemental material}

\section{Code}

We provide the code for this work at \url{https://github.com/StijnVerdenius/Lat-PFN}.

\section{Broader Societal Impacts}
\label{sec:appendix:impact}

It is our belief that the work presented in this paper does not pose major negative societal risks. The model is trained exclusively on synthetic time series data, thus eliminating any risk related to data privacy, disinformation or consent. On the contrary, given how ubiquitous time series are in the real world, we hope that our model for zero-shot forecasting will provide a net benefit for society, for a few different reasons. Firstly, it is significantly more efficient to train than many of the baselines we evaluate against, as pointed out in Section~\ref{sec:experimental-details}. Secondly, as a zero-shot foundation model, it only requires training once, before being used out-of-the-box. Thirdly, many sectors in society could benefit from having improved forecasting abilities for assisting day-to-day decision-making. We hope this will result in widespread adoption of pre-trained models for forecasting tasks, thus improving industry effectiveness and reducing energy consumption and CO2 emissions. % Consequently, we believe that safeguards are not applicable at this time.  

\section{Derivation PFN NLLLoss}
\label{sec:appendix:pfn}

This derivation, adapted from \cite{muller2021transformers}, illustrates the use of a cross-entropy loss with a synthetic prior as an approximation of the Predictive Posterior Distribution (PPD). This is a more effective way to learn Bayesian inference, as opposed to estimating the true posterior, which is often intractable. Furthermore, it yields a flexible approach by encoding expert data, which makes it highly adaptable for developing downstream applications \citep{muller2021transformers}.

Assume we want to approximate PPD \(P(y|x,\mathcal{D})\) with network \(Q(y|x,\mathcal{D})\), for contexts \(\mathcal{D}\). We start with the expectation over the contexts, of the KL-divergence between distributions \(P(.)\) and \(Q(.)\):
\begin{align*}
    \displaystyle
   && &\mathbb{E}_{x,\mathcal{D}} \Bigg[ \quad KL\big( P(y|x,\mathcal{D}), Q_\theta(y|x,\mathcal{D})     \Big) &&\Bigg] & & \\
    &&= \quad &\mathbb{E}_{x,\mathcal{D}} \Bigg[ \: - \int_y   P(y|x,\mathcal{D}) \log \frac{Q_\theta(y|x,\mathcal{D})}{P(y|x,\mathcal{D})} &&\Bigg] & & \\
    &&= \quad &\mathbb{E}_{x,\mathcal{D}} \Bigg[ \: - \int_y   P(y|x,\mathcal{D}) \log Q_\theta(y|x,\mathcal{D}) &&\Bigg] \: &&+ \:  \mathbb{E}_{x,\mathcal{D}} \Bigg[ \: \int_y   P(y|x,\mathcal{D}) \log P(y|x,\mathcal{D}) \Bigg] &\\
    &&= \quad &\mathbb{E}_{x,\mathcal{D}} \Bigg[ \quad \bm{H}\big(P(y|x,\mathcal{D}), Q_\theta(y|x,\mathcal{D})     \Big) &&\Bigg]  \: &&+ \: C &
\end{align*}
We find equivalence to the entropy function \(\bm{H}(.)\) and constant \(C\). The latter in independent of parameters \(\theta\) and is dropped in optimization \citep{muller2021transformers}.  Next, we elect to incorporate the PFN synthetic prior simulation by sampling data from our simulation: 

\begin{align*}
    \displaystyle
    &&\mathcal{L} \quad &= \quad \mathbb{E}_{\{(x,y)\cup\mathcal{D}\}\sim S(. | \psi)} \Bigg[ \quad\quad \bm{H}\big(P(y|x,\mathcal{D}), Q_\theta(y|x,\mathcal{D})     \Big) \quad &&\Bigg] &\\
    && \quad &= \quad \mathbb{E}_{\{(x,y)\cup\mathcal{D}\}\sim S(. | \psi)} \Bigg[ \quad  -  \sum_{y}  \:  P(y|x,\mathcal{D}) \log Q_\theta(y|x,\mathcal{D})    \Big) \quad &&\Bigg] &
\end{align*}

These formulas show that, for the purpose of optimization, this is equivalent to applying the cross-entropy loss -- a commonly used loss function. The output space is split in bins to allow the use of a cross-entropy loss for regression and forecasting problems -- following the PFN example \citep{muller2021transformers}. Furthermore, the synthetic prior provides control over the prior distribution of contexts and access to infinite training data resources, including contexts, held-out examples, and targets \citep{muller2021transformers}.

Finally, in this work, we introduce a separation of concerns between predicting and decoding. The end-to-end PPD task is effectively split into (a) learning a good way of summarizing the prior distribution within the latent prediction distribution \(P(\hat{\bar{y}} | x, \mathcal{D}, \theta)\), and then (b) separately learning an approximated posterior based on that summary \(P(y | \hat{\bar{y}}, \theta )\) -- again without assumptions on output distribution family. Although the optimization is decoupled, there remains a causal relationship between the output distribution over \(y\) and it dependent variables: \(x\), \(\mathcal{D}\), and parameters \(\theta\). Therefore, we argue that this approach is a reasonable approximation of the PPD -- given the latent predictions as an intermediary.

\FloatBarrier

\section{Reproducibility Details}
\label{sec:appendix:reproducibility}

This section is dedicated to providing comprehensive reproducibility details for the research presented in this document. By including this appendix, we aim to uphold transparency standards in research.

\subsection{Exhaustive Description Synthesis Training Data }
\label{sec:appendix:reproducibility:simulation}

Following the example set by \cite{dooley2023forecastpfn} we define a synthetic prior that leverages the underlying components of time series data, namely trend, seasonality, and noise. The trend parameter is made up of linear and exponential components; the seasonality parameter is made up of annual, monthly, and weekly components and the noise is derived from a Weibull distribution:

\[
    y_t = \psi(t) \cdot z_t = trend(t) \cdot seasonality(t) \cdot z_t
\]

The linear and exponential trend components are defined by two constituent parts each: scaler and offset. The scaler $m$ is a context-wide parameter, therefore we use triple sampling to generate them. The offset $c$ is domain-specific, so we directly sample them, uniformly, from a range of hyperpriors. The formula for the trend component is the same as defined by  \citep{dooley2023forecastpfn}: 

\[
    trend(t) = 1 + (m_{lin} \cdot t + c_{lin}) \cdot (m_{exp} \cdot c_{t}^{exp})
\]

The seasonality parameters are all context-dependant, therefore we apply triple sampling to derive them. Additionally, we introduce a new parameter, \(\delta\) to define the number of frequency features for each seasonality parameter. We use triple sampling for the seasonality parameters and direct univariate sampling for the  \(\delta\) parameter. The formulas have once again been adapted from \citep{dooley2023forecastpfn}:

\begin{align*}
    \displaystyle
    &seasonal(t) = seasonal_{week}(t) \cdot seasonal_{month}(t) \cdot seasonal_{year}(t) \\\\
    &\delta_v \sim \mathcal{U}(\alpha^*_\delta) \\
    &seasonal_v(t) = 1 + m_v \sum_{f=1}^{\delta_v}\left[ c_{f, v}\sin\left({2f\pi\frac{t}{p_v}}\right) + d_{f, v}\cos\left({2f\pi\frac{t}{p_v}}\right) \right] \\\\
    &v \in{\{week, month, year\}} \\\\
    &p_{week} = 7, ~ p_{month} = 30.417, ~ p_{year} = 30.417 \\\\
    &c_{v} \sim \mathcal{N}\left(0, \frac{1}{\lvert \delta_v \rvert}\right), ~ d_{v} \sim \mathcal{N}\left(0, \frac{1}{\lvert \delta_v \rvert}\right)\\
\end{align*}

The noise component follows the definition laid out by \citep{dooley2023forecastpfn}, the only difference being that we sample a single \(k\) parameter per context, and use it in the Weibull distribution:
\begin{align*}
    &z = 1 + m_{noise} (z - \overline{z})\\
    &z \sim Weibull(1, k), ~ \overline{z} = \left(\ln{2}\right)^{1/k}, ~ m_{noise} \in{\{ \mathcal{U}(0, 0.1), \mathcal{U}(0.2, 0.4), \mathcal{U}(0.6, 0.8)\}}
\end{align*}

Unlike \cite{dooley2023forecastpfn}, we train using an normalized time dimension. In order to include many different types of temporal patterns, we define a resolution parameter \(\rho\), which determines how many points are represented per interval, for each individual time series. We define this as the relation between the sequence dimension and resolution parameter:
\[
    t = \frac{S}{\rho} \\
\]

Finally, for the resolution and exponential-trend parameters, we obtain the context parameters \(\alpha_c\), described in Section~\ref{sec:methodology:simulation}, by sampling log-uniformly, due to the logarithmic nature of these components. Specifically, we guarantee a median sample value without having a forced symmetric range around that median value. To do so, we (a) define a mapping to log space, (b) sample uniformly, and (c) inverse the mapping after. The formula is as follows:
\begin{align*}
\displaystyle 
map(x, \kappa) \quad &= \quad log_2(x \cdot \kappa + 1) \\
map^{-1}(x, \kappa) \quad &= \quad  \frac{2^x - 1}{\kappa} \\
\alpha_{c,mapped} \quad &\sim \quad  \mathcal{U} \Big( map(\alpha^*, \kappa) \Big) \\
\alpha_c \quad &= \quad map^{-1}(\alpha_{c,mapped}, \kappa)
\end{align*}
For multiplier \(\kappa\) and function \(map(.)\).  Please see Table~\ref{tab:hyperpriors} for details on the multipliers and hyperpriors.

% map(\alpha^{}_c) &\sim \mathcal{U} \Big( \log_2(\alpha^*_{\text{min}} \kappa + 1), \log_2(\alpha^*_\text{max}\kappa + 1) \Big)  \\
% \displaystyle \alpha_c &= \frac{2^{map(\alpha_c)}}{\kappa} -1
\begin{table}[h]
\centering
\caption{Hyperpriors for Simulation Engine}
\label{tab:hyperpriors}
\begin{tabular}{@{}ccc@{}}
\toprule
\textbf{Hyperparameter} & \textbf{Min Value} & \textbf{Max Value} \\ 
\midrule
\textbf{Seasonality}    &                    &                    \\
Annual Frequency Scale        & \(-8.0\)             & \(8.0\)              \\
Monthly Frequency Scale       & \(-4.0\)             & \(4.0\)              \\
Weekly Frequency Scale        & \(-2.0\)             & \(2.0\)              \\
Variance           & \multicolumn{2}{c}{0.15}  \\ 
\midrule
\textbf{Trend}         &                    &                    \\
Linear       & \(-0.015\)           & \(0.015\)            \\
Linear Variance & \multicolumn{2}{c}{0.005}  \\
Exponential  & \(0.996\)            & \(1.0016\)           \\
Exponential  Variance  & \multicolumn{2}{c}{0.001} \\
Exponential Multiplier  \(\kappa_{m_{exp}}\)               & \multicolumn{2}{c}{507}    \\ 
\midrule
\textbf{Noise}         &                    &                    \\
Noise (\(k\))                          & \(0.8\)              & \(5\)                \\ 
\midrule
\textbf{Resolution}    &                    &                    \\
Resolution Min                         & \(0.1\)              & \(1.0\)              \\
Resolution Multiplier \(\kappa_\rho\)                 & \multicolumn{2}{c}{53.6}   \\ 
\midrule
\textbf{Offset}        &                    &                    \\
Linear Offset                          & \(-1\)               & \(2\)                \\
Exponential Offset                     & \(-1\)               & \(2\)                \\ 
\midrule
\textbf{Harmonics}     &                    &                    \\
\(\delta_{\text{harmonics}}\)          & \(4\)                & \(12\)               \\ 
\bottomrule
\end{tabular}
\end{table}

\FloatBarrier

\subsection{Baselines}

\paragraph{Forecasting} For forecasting we use the following baselines:
\begin{itemize}
    \item FBProphet \citep{taylor2018forecasting}, a common industry standard with respect to the type of univariate series we model. We use the publicly available repository \citep{FBProphet} and manually tune the parameter \texttt{changepoint\_prior\_scale=1.0}. 
    \item ARIMA \citep{contreras2003arima}, another classic baseline from the industry. We use the Statsmodel implementation \citep{statsmodelsARIMA} publicly available and pass the argument \texttt{order=(5, 1, 0)}. 
    \item ForecastPFN \citep{dooley2023forecastpfn} is a flexible in-context forecasting model and one of the inspirations for this work. We download their open-source repository available under Apache Licence 2.0 and trained model weights. We adopt their inference script to work in our repository. We have tried several data normalizations by hand to get the most out their model and picked the best one, which turned out to be Z-normalization with 2std -- on top of the normalizations present in the repository already.
\end{itemize}

\paragraph{Classification} For classification we use the TS2Vec baseline \citep{yue2022ts2vec}, as it is a strong multi-purpose representation learner and also an inspiration for this work, although TS2Vec does not claim to implement zero-shot forecasting. They open-source their training code under the MIT Licence, so we train TS2Vec from scratch on the train splits of the UCR \citep{UCRArchive2018} archive, after which we evaluate on the corresponding test splits with the resulting frozen TS2Vec architecture

\subsection{Benchmark Preprocessing}

We use five datasets for our experiments, detailed in section \ref{sec:experimental-details}. These are considered standard datasets for benchmarking the performance of time series forecasting models \cite{oreshkin2021meta, wu2021autoformer, zhou2021informer}. 

For each dataset, we create univariate time series out of each of the columns. Additionally, we further split up the univariate series along the time dimension to create single entities of different granularity, which we then use for context creation. We normalize the time dimension of each dataset so that it fits within the normalized \([-3,1]\) interval. Furthermore, we apply Z-score normalization, with 2 standard deviations, to the values of each univariate time series.

The specifics of each dataset are presented in the following sections. We reference the original sources of the datasets. However, practically we obtained the actual files from the ForecastPFN repository \citep{dooley2023forecastpfn, Abacusai} 

\subsubsection{Illness}
Illness is a dataset of influenza-like illness patients in the United States. It reports patients data in a weekly time granularity, from 1997 to 2024. \cite{Centers_for_Disease_Control_and_Prevention}

During pre-processing, we create files for each column of the dataset and split them along the time dimension, with a monthly periodicity.

We define a sequence length of 160 equidistant intervals, of which the last 25\% are used as the prediction target. We use a rolling window over the dataset, with a stride of 1. 

We hand-picked 8 windows of the historic series as context examples and kept 1 as held-out starting from forecast date 01-01-2018, which we evaluate the zero-shot performance on.\\

\begin{center}
\begin{tabular}{ll}
 Context & Held-out\\  \midrule
 AGE\_65 2008-2012 &  ILITOTAL 2018-2022 \\
 AGE\_65 2013-2017 &   \\
 AGE\_0-4 2008-2012 & \\
 AGE\_0-4 2013-2017 & \\
  AGE\_5-24 2008-2012 & \\
 AGE\_5-24 2013-2017 & \\
 TOTAL\_PATIENTS 2008-2012 & \\
 TOTAL\_PATIENTS 2013-2017
\end{tabular}
\end{center}

The illness dataset is provided by a public government institution (CDC), which does not provide an explicit license.

\subsubsection{EttH1 \& EttH2}
The  ETT (Electricity Transformer Temperature) dataset is an electricity power deployment dataset. The data is reported both hourly and minute-wise, from 2016-07-01 to 2018-06-26. The dataset reports two different series, ETT1, and ETT2, corresponding to two different transformer stations, where the data was recorded \cite{zhou2021informer}.

We experiment on the hourly series but aggregate the data monthly and create univariate time series out of each column. Following the dataset's original directions, we use the OT column as held-out, on which we evaluate the zero-shot performance of the model.

We define a sequence length of 240 equidistant intervals, of which the last 25\% are used as targets. We use a rolling window over each time series with a stride of 10.

We run 2 separate experiments for this dataset, one for each transformer station series. We change both the held-out series and context for each experiment.
\begin{itemize}
    \item Experiment 1: ETTh1\\
    We provide a context dimension of 14 and leave one held-out time series.\\
        \begin{center}
            \begin{tabular}{ll}
             Context & Held-out\\  \midrule
             ETTh1 HUFL 2017/01 - 2017/05 &  ETTh1 OT 2018/01 - 2018/05 \\
             ETTh1 HULL 2017/01 - 2017/05 &   \\
             ETTh1 LUFL 2017/01 - 2017/05 & \\
             ETTh1 LULL 2017/01 - 2017/05 & \\
             ETTh1 MUFL 2017/01 - 2017/05 & \\
             ETTh1 MULL 2017/01 - 2017/05 & \\
             ETTh2 HULL 2017/01 - 2017/05 &   \\
             ETTh2 LUFL 2017/01 - 2017/05 & \\
             ETTh2 LULL 2017/01 - 2017/05 & \\
             ETTh2 MUFL 2017/01 - 2017/05 & \\
             ETTh2 MULL 2017/01 - 2017/05 & \\
            \end{tabular}
    \end{center}

    \item Experiment 2 - ETTh2:\\
    We provide a context dimension of 4 and leave one held-out time series.\\

    \begin{center}
            \begin{tabular}{ll}
             Context & Held-out\\  \midrule
             ETTh1 HUFL 2017/01 - 2017/05 &  ETTh2 OT 2018/01 - 2018/05 \\
             ETTh1 OT 2017/01 - 2017/05 & \\
             ETTh2 HUFL 2017/01 - 2017/05 & \\
             ETTh2 HULL 2017/01 - 2017/05
            \end{tabular}
    \end{center}
    
\end{itemize}

Both datasets are under Creative Commons Attribution-NoDerivatives 4.0 International.

\subsubsection{Traffic}
Traffic is a dataset reporting freeway congestion in California, USA. We used the data from \cite{dooley2023forecastpfn}, who originally sourced it from \cite{pems2021caltrans}. The data is aggregated daily and consists of 860 numerical columns (named 1-860), plus a target column (OT).

We aggregate the data monthly and create univariate time series for each of the columns.

We define a sequence length of 240 equidistant intervals, of which the last 25\% are used as targets. We use a rolling window over each time series with a stride of 1.

We provide a context dimension of 8 and leave two held-out time series.\\

\begin{center}
    \begin{tabular}{ll}
     Context & Held-out\\  \midrule
     120 2016/11 - 2017/05 &  360 2017/11 - 2018/05 \\
     120 2016/10 - 2017/04 &  360 2017/08 - 2018/02 \\
     293 2016/11 - 2017/05 & \\
     293 2016/10 - 2017/04 & \\
     710 2016/11 - 2017/05 & \\
     710 2016/10 - 2017/04 & \\
     405 2016/11 - 2017/05 & \\
     405 2016/10 - 2017/04
    \end{tabular}
\end{center}

\subsubsection{ECL}
ECL (Electricity Consuming Load) \cite{misc_electricityloaddiagrams20112014_321} reports electricity consumption in Kwh with an hourly frequency from 2016-07-01 to 2019-07-02. The data consists of 320 numerical columns (named 0-319), plus a target column (OT)

We aggregate the data daily and create univariate time series for each of the columns.

We define a sequence length of 240 equidistant intervals, of which the last 25\% are used as targets. We use a rolling window over each time series with a stride of 10.

We provide a context dimension of 30 and leave two held-out time series.\\

        \begin{center}
            \begin{tabular}{ll}
             Context & Held-out\\  \midrule
             208 2018/04/01 - 2018/04/14 &  OT 2019/05/01 - 2018/05/14 \\
             208 2018/04/07 - 2018/04/20 &  OT 2019/04/01 - 2018/04/14 \\
             208 2018/04/17 - 2018/04/30 & \\
             208 2017/04/01 - 2017/04/14 & \\
             208 2017/04/07 - 2017/04/20 & \\
             208 2017/04/17 - 2017/04/30 & \\
             208 2019/01/01 - 2019/01/14 & \\
             208 2019/01/07 - 2019/01/20 & \\
             208 2019/01/17 - 2019/01/30 & \\
             208 2019/02/01 - 2019/02/14 & \\
             208 2019/02/07 - 2019/02/20 & \\
             208 2019/02/17 - 2019/03/02 & \\
             208 2019/03/01 - 2019/03/14 & \\
             208 2019/03/07 - 2019/03/20 & \\
             208 2019/03/17 - 2019/03/30 & \\
             313 2018/04/01 - 2018/04/14 & \\
             313 2018/04/07 - 2018/04/20 & \\
             313 2018/04/17 - 2018/04/30 & \\
             313 2017/04/01 - 2017/04/14 & \\
             313 2017/04/07 - 2017/04/20 & \\
             313 2017/04/17 - 2017/04/30 & \\
             313 2019/01/01 - 2019/01/14 & \\
             313 2019/01/07 - 2019/01/20 & \\
             313 2019/01/17 - 2019/01/30 & \\
             313 2019/02/01 - 2019/02/14 & \\
             313 2019/02/07 - 2019/02/20 & \\
             313 2019/02/17 - 2019/03/02 & \\
             313 2019/03/01 - 2019/03/14 & \\
             313 2019/03/07 - 2019/03/20 & \\
             313 2019/03/17 - 2019/03/30 & \\

            \end{tabular}
    \end{center}

This dataset is under CC BY 4.0 licence.

\subsection{Training \& Tuning}
\label{sec:appendix:reproducibility:training}
Please consider the following training and tuning details:
\begin{itemize}
    \item We use a Cosine Annealing With Decay and Warm restarts Learning rate schedule. Base learning rate is 9e-4 and the scheduler parameters consist of a decay of 0.96 and a T0 of 9.
    \item We train with a linear warmup of 95 epochs for JEPA-EMA decay and weight decay. Increasing the former from 0.9952 to 1.0 and the latter from 1.77e-4 to 4.9e-2.
    \item We tune all of the above optimizer hyperparameters, \(\lambda_\text{di}\) and \(\lambda_\text{latent}\) with Optuna \citep{optuna_2019} in the 3 runs, spanning a total of 216 hours. 
    \item We apply MUP \citep{yang2022tensor}, which allows us to tune a smaller model and zero-shot transfer the hyperparameters we found. This also proved essential for training stability. MUP also provides us with the MUP-AdamW optimizer \citep{yang2022tensor, loshchilov2017decoupled, kingma2014adam}, which was the optimizer in all experiments. 
    \item All the variables in the bullets below were not tuned but hand-picked by either trial and error or expert knowledge. 
    \item We apply TF32 quantization but no mixed-precision training since this gave stability issues.
    \item For the final model we train with a batch size of 32, a context size of 14, a history of 180, horizon of 60 and held-out size of 2.
    \item For system identification we multi-target regress to 10 simulation parameters, unit-normalized. Specifically, these variables and normalization scales are:
    \begin{itemize}
        \item Annual frequency scale: unit-normalized
        \item Monthly frequency scale unit-normalized
        \item Weekly frequency scale: unit-normalized
        \item Trend linear: unit-normalized
        \item Trend exponential: log-unit-normalized
        \item Offset linear: unit-normalized
        \item Offset exponential: unit-normalized
        \item Noise scale: log-unit-normalized
        \item Resolution: log-unit-normalized
    \end{itemize}
    \item Since we have a synthetic dataset, we have no natural end of an epoch. We elect to choose a 250-batch epoch before updating schedules.
    \item We map our training data to an normalized time domain of \([-3, 1]\), with the forecast moment at the origin.
    \item We define 100 bins on the normalized \([-3.5,3.5]\) range, as the output space for the decoder.
    \item Model-width is set at 512 neurons for the final model and 128 neurons for tuning. Corresponding depths are:
    \begin{itemize}
        \item Embedder: 8 layers
        \item Predictor: 3 layers
        \item SI-head: 2 layers
        \item Decoder: 3 layers
    \end{itemize}
    \item On top of the compute documented in Section~\ref{sec:experimental-details}, which indicates the compute required to train one seed of the final model, we additionally used more compute for experiments in the discovery phase. How much is unfortunately not something we have kept track of.
\end{itemize}

\newpage
\FloatBarrier

\begin{figure}[h!]
    \centering
    \includegraphics[width=0.25\linewidth]{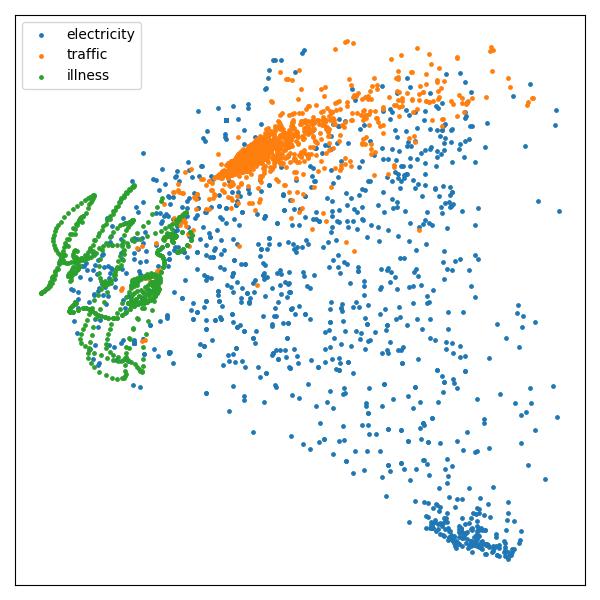}
    \caption{PCA plot of fixed-length summary embeddings grouped by domain}
    \label{fig:appendix:pca}    
\end{figure}

% \begin{figure}[p]
\section{Latent Space Visualised}
\label{sec:appendix:latent_space}
\vspace{3cm}

% \vfill
%     \centering
%     \includegraphics[width=0.25\linewidth]{figures/pca_42.jpg}
%     \caption{PCA plot of fixed-length summary embeddings grouped by domain}
%     \label{fig:appendix:pca}
% \end{figure}

\begin{figure}[h!]
  \centering
  \captionsetup[subfigure]{justification=centering, labelformat=empty}
\subfloat[\textit{\footnotesize ETTh1}]{
  \includegraphics[width=0.8\linewidth]{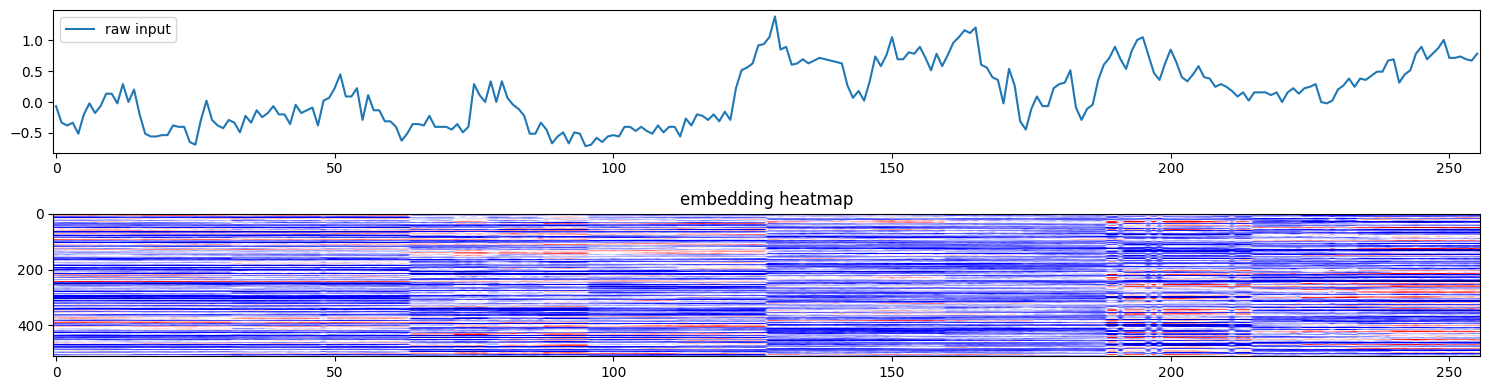}\label{fig:appendix:ts-tokens:etth1}
} \\
\subfloat[\textit{\footnotesize ETTh2}]{
  \includegraphics[width=0.8\linewidth]{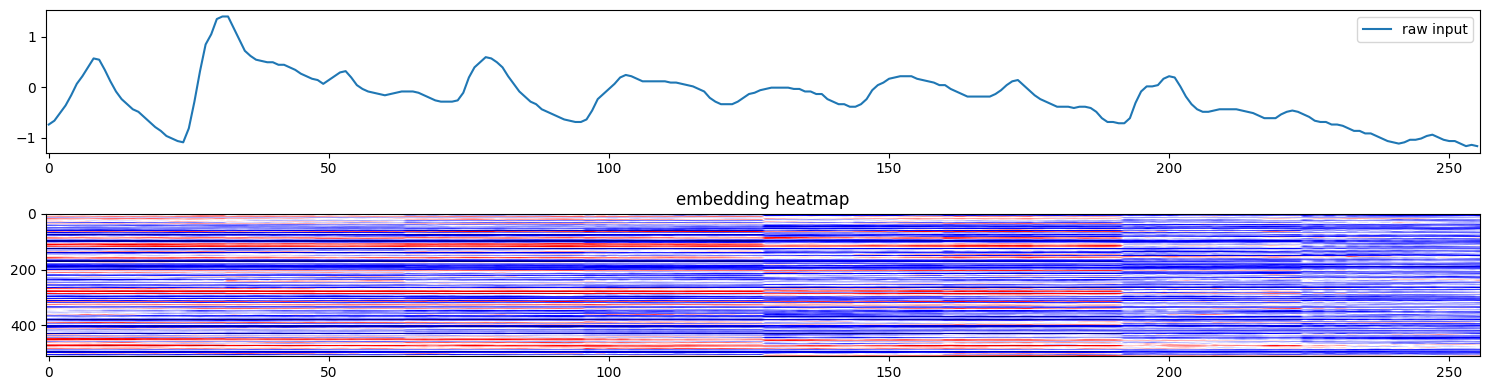}\label{fig:appendix:ts-tokens:etth2}
} \\
\subfloat[\textit{\footnotesize Traffic }]{
  \includegraphics[width=0.8\linewidth]{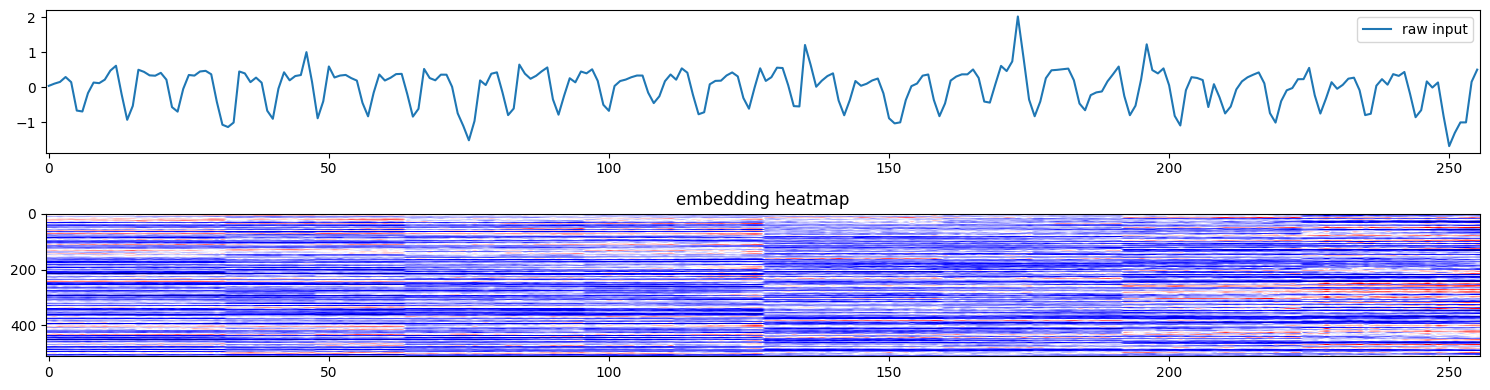} \label{fig:appendix:ts-tokens:traffic}
}
  \caption{Additional per-step time series embeddings}
  \label{fig:appendix:ts-tokens}
\end{figure} 

\FloatBarrier

\section{Additional results}
\label{sec:appendix:results}

\FloatBarrier

\FloatBarrier

\subsection{UCR \citep{UCRArchive2018} Accuracy by Dataset}

% \usepackage{booktabs}
\begin{table}[h!]
\centering
\caption{UCR scores per dataset: A-F}
\label{tab:classification-per-ds:a-f}
\begin{tabular}{@{}lrlrl@{}}
\toprule
\textbf{UCR Datasets}                & \multicolumn{2}{l}{\textbf{Our work}} & \multicolumn{2}{l}{\textbf{TS2Vec}}  \\ \midrule
Acsf1 Test & 36.00 \%&{\scriptsize\(\pm\) 5.0 \% } & \textbf{43.00} \%&{\scriptsize\(\pm\) 2.8 \% } \\
Adiac Test & \textbf{03.07} \%&{\scriptsize\(\pm\) 2.0 \% } & 02.05 \%&{\scriptsize\(\pm\) 0.0 \% } \\
Allgesturewiimotex Test & \textbf{24.50} \%&{\scriptsize\(\pm\) 5.8 \% } & 21.21 \%&{\scriptsize\(\pm\) 5.6 \% } \\
Allgesturewiimotey Test & \textbf{24.39} \%&{\scriptsize\(\pm\) 5.0 \% } & 16.36 \%&{\scriptsize\(\pm\) 2.1 \% } \\
Allgesturewiimotez Test & \textbf{19.25} \%&{\scriptsize\(\pm\) 4.9 \% } & 17.50 \%&{\scriptsize\(\pm\) 1.5 \% } \\
Arrowhead Test & 43.00 \%&{\scriptsize\(\pm\) 10.2 \% } & \textbf{44.86} \%&{\scriptsize\(\pm\) 4.4 \% } \\
Bme Test & \textbf{51.83} \%&{\scriptsize\(\pm\) 2.0 \% } & 49.00 \%&{\scriptsize\(\pm\) 3.3 \% } \\
Beef Test & \textbf{41.67} \%&{\scriptsize\(\pm\) 8.8 \% } & 38.33 \%&{\scriptsize\(\pm\) 2.4 \% } \\
Beetlefly Test & 56.25 \%&{\scriptsize\(\pm\) 4.8 \% } & \textbf{75.00} \%&{\scriptsize\(\pm\) 14.1 \% } \\
Birdchicken Test & 67.50 \%&{\scriptsize\(\pm\) 15.5 \% } & \textbf{70.00} \%&{\scriptsize\(\pm\) 7.1 \% } \\
Cbf Test & \textbf{47.92} \%&{\scriptsize\(\pm\) 10.8 \% } & 40.33 \%&{\scriptsize\(\pm\) 10.2 \% } \\
Car Test & \textbf{21.67} \%&{\scriptsize\(\pm\) 0.0 \% } & \textbf{21.67} \%&{\scriptsize\(\pm\) 0.0 \% } \\
Chinatown Test & \textbf{79.23} \%&{\scriptsize\(\pm\) 15.2 \% } & 78.57 \%&{\scriptsize\(\pm\) 3.1 \% } \\
Chlorineconcentration Test & \textbf{53.65} \%&{\scriptsize\(\pm\) 0.6 \% } & 53.26 \%&{\scriptsize\(\pm\) 0.0 \% } \\
Cincecgtorso Test & \textbf{25.85} \%&{\scriptsize\(\pm\) 2.1 \% } & 25.18 \%&{\scriptsize\(\pm\) 0.6 \% } \\
Coffee Test & 72.32 \%&{\scriptsize\(\pm\) 5.4 \% } & \textbf{82.14} \%&{\scriptsize\(\pm\) 0.0 \% } \\
Computers Test & 53.90 \%&{\scriptsize\(\pm\) 2.5 \% } & \textbf{66.80} \%&{\scriptsize\(\pm\) 0.6 \% } \\
Cricketx Test & \textbf{21.15} \%&{\scriptsize\(\pm\) 8.7 \% } & 13.85 \%&{\scriptsize\(\pm\) 0.0 \% } \\
Crickety Test & \textbf{21.47} \%&{\scriptsize\(\pm\) 10.9 \% } & 16.54 \%&{\scriptsize\(\pm\) 1.3 \% } \\
Cricketz Test & \textbf{20.45} \%&{\scriptsize\(\pm\) 10.4 \% } & 13.85 \%&{\scriptsize\(\pm\) 0.7 \% } \\
Crop Test & 31.78 \%&{\scriptsize\(\pm\) 6.2 \% } & \textbf{32.07} \%&{\scriptsize\(\pm\) 0.3 \% } \\
Diatomsizereduction Test & \textbf{31.70} \%&{\scriptsize\(\pm\) 3.3 \% } & 30.07 \%&{\scriptsize\(\pm\) 0.0 \% } \\
Distalphalanxoutlineagegroup Test & \textbf{63.49} \%&{\scriptsize\(\pm\) 8.3 \% } & 46.76 \%&{\scriptsize\(\pm\) 0.0 \% } \\
Distalphalanxoutlinecorrect Test & \textbf{58.33} \%&{\scriptsize\(\pm\) 0.0 \% } & \textbf{58.33} \%&{\scriptsize\(\pm\) 0.0 \% } \\
Distalphalanxtw Test & \textbf{59.35} \%&{\scriptsize\(\pm\) 3.7 \% } & 30.22 \%&{\scriptsize\(\pm\) 0.0 \% } \\
Dodgerloopday Test & 26.88 \%&{\scriptsize\(\pm\) 5.2 \% } & \textbf{38.12} \%&{\scriptsize\(\pm\) 0.9 \% } \\
Dodgerloopgame Test & \textbf{60.87} \%&{\scriptsize\(\pm\) 11.9 \% } & 48.55 \%&{\scriptsize\(\pm\) 2.0 \% } \\
Dodgerloopweekend Test & 72.10 \%&{\scriptsize\(\pm\) 26.6 \% } & \textbf{90.22} \%&{\scriptsize\(\pm\) 1.5 \% } \\
Ecg200 Test & \textbf{64.00} \%&{\scriptsize\(\pm\) 0.0 \% } & \textbf{64.00} \%&{\scriptsize\(\pm\) 0.0 \% } \\
Ecg5000 Test & 86.03 \%&{\scriptsize\(\pm\) 4.0 \% } & \textbf{86.33} \%&{\scriptsize\(\pm\) 1.7 \% } \\
Ecgfivedays Test & \textbf{52.41} \%&{\scriptsize\(\pm\) 5.4 \% } & 49.71 \%&{\scriptsize\(\pm\) 0.0 \% } \\
Eoghorizontalsignal Test & 16.99 \%&{\scriptsize\(\pm\) 2.8 \% } & \textbf{29.42} \%&{\scriptsize\(\pm\) 0.6 \% } \\
Eogverticalsignal Test & 16.44 \%&{\scriptsize\(\pm\) 5.2 \% } & \textbf{21.13} \%&{\scriptsize\(\pm\) 0.2 \% } \\
Earthquakes Test & \textbf{74.82} \%&{\scriptsize\(\pm\) 0.0 \% } & \textbf{74.82} \%&{\scriptsize\(\pm\) 0.0 \% } \\
Electricdevices Test & \textbf{52.58} \%&{\scriptsize\(\pm\) 3.0 \% } & 52.41 \%&{\scriptsize\(\pm\) 2.7 \% } \\
Ethanollevel Test & 25.95 \%&{\scriptsize\(\pm\) 0.6 \% } & \textbf{26.30} \%&{\scriptsize\(\pm\) 0.1 \% } \\
Faceall Test & 25.72 \%&{\scriptsize\(\pm\) 4.5 \% } & \textbf{41.57} \%&{\scriptsize\(\pm\) 1.2 \% } \\
Facefour Test & \textbf{38.92} \%&{\scriptsize\(\pm\) 10.1 \% } & 27.27 \%&{\scriptsize\(\pm\) 12.9 \% } \\
Facesucr Test & \textbf{23.33} \%&{\scriptsize\(\pm\) 3.8 \% } & 16.32 \%&{\scriptsize\(\pm\) 2.8 \% } \\
Fiftywords Test & \textbf{22.64} \%&{\scriptsize\(\pm\) 5.6 \% } & 12.53 \%&{\scriptsize\(\pm\) 0.0 \% } \\
Fish Test & 21.00 \%&{\scriptsize\(\pm\) 6.1 \% } & \textbf{22.57} \%&{\scriptsize\(\pm\) 0.4 \% } \\
Forda Test & \textbf{67.42} \%&{\scriptsize\(\pm\) 10.8 \% } & 62.88 \%&{\scriptsize\(\pm\) 1.9 \% } \\
Fordb Test & \textbf{57.69} \%&{\scriptsize\(\pm\) 6.4 \% } & 56.11 \%&{\scriptsize\(\pm\) 3.1 \% } \\
Freezerregulartrain Test & \textbf{76.06} \%&{\scriptsize\(\pm\) 0.9 \% } & 75.79 \%&{\scriptsize\(\pm\) 0.0 \% } \\
Freezersmalltrain Test & 73.27 \%&{\scriptsize\(\pm\) 5.2 \% } & \textbf{75.86} \%&{\scriptsize\(\pm\) 0.0 \% } \\
Fungi Test & \textbf{67.74} \%&{\scriptsize\(\pm\) 10.2 \% } & 46.24 \%&{\scriptsize\(\pm\) 0.0 \% } \\ \bottomrule
\end{tabular}
\end{table}

\begin{table}[h!]
\centering
\caption{UCR scores per dataset: G-R}
\label{tab:classification-per-ds:g-r}
\begin{tabular}{@{}lrlrl@{}}
\toprule
\textbf{UCR Datasets}                & \multicolumn{2}{l}{\textbf{Our work}} & \multicolumn{2}{l}{\textbf{TS2Vec}}  \\ \midrule

Gesturemidaird1 Test & \textbf{25.77} \%&{\scriptsize\(\pm\) 6.6 \% } & 23.08 \%&{\scriptsize\(\pm\) 4.4 \% } \\
Gesturemidaird2 Test & \textbf{30.00} \%&{\scriptsize\(\pm\) 7.2 \% } & 21.15 \%&{\scriptsize\(\pm\) 0.5 \% } \\
Gesturemidaird3 Test & \textbf{14.81} \%&{\scriptsize\(\pm\) 6.9 \% } & 08.46 \%&{\scriptsize\(\pm\) 2.2 \% } \\
Gesturepebblez1 Test & 25.44 \%&{\scriptsize\(\pm\) 11.9 \% } & \textbf{45.64} \%&{\scriptsize\(\pm\) 7.0 \% } \\
Gesturepebblez2 Test & \textbf{27.37} \%&{\scriptsize\(\pm\) 13.6 \% } & 26.27 \%&{\scriptsize\(\pm\) 6.7 \% } \\
Gunpointagespan Test & 52.53 \%&{\scriptsize\(\pm\) 3.8 \% } & \textbf{54.43} \%&{\scriptsize\(\pm\) 4.9 \% } \\
Gunpointmaleversusfemale Test & 55.06 \%&{\scriptsize\(\pm\) 5.1 \% } & \textbf{84.49} \%&{\scriptsize\(\pm\) 0.4 \% } \\
Gunpointoldversusyoung Test & 52.38 \%&{\scriptsize\(\pm\) 0.0 \% } & \textbf{100.00} \%&{\scriptsize\(\pm\) 0.0 \% } \\
Gunpoint Test & 54.50 \%&{\scriptsize\(\pm\) 7.5 \% } & \textbf{57.00} \%&{\scriptsize\(\pm\) 10.8 \% } \\
Ham Test & \textbf{58.81} \%&{\scriptsize\(\pm\) 6.7 \% } & 51.43 \%&{\scriptsize\(\pm\) 0.0 \% } \\
Handoutlines Test & \textbf{64.05} \%&{\scriptsize\(\pm\) 0.0 \% } & \textbf{64.05} \%&{\scriptsize\(\pm\) 0.0 \% } \\
Haptics Test & \textbf{21.19} \%&{\scriptsize\(\pm\) 0.8 \% } & 20.78 \%&{\scriptsize\(\pm\) 0.0 \% } \\
Herring Test & \textbf{59.38} \%&{\scriptsize\(\pm\) 0.0 \% } & \textbf{59.38} \%&{\scriptsize\(\pm\) 0.0 \% } \\
Housetwenty Test & 62.18 \%&{\scriptsize\(\pm\) 5.9 \% } & \textbf{71.01} \%&{\scriptsize\(\pm\) 0.6 \% } \\
Inlineskate Test & 17.68 \%&{\scriptsize\(\pm\) 0.7 \% } & \textbf{18.00} \%&{\scriptsize\(\pm\) 1.0 \% } \\
Insectepgregulartrain Test & 85.84 \%&{\scriptsize\(\pm\) 25.7 \% } & \textbf{100.00} \%&{\scriptsize\(\pm\) 0.0 \% } \\
Insectepgsmalltrain Test & 76.00 \%&{\scriptsize\(\pm\) 24.5 \% } & \textbf{100.00} \%&{\scriptsize\(\pm\) 0.0 \% } \\
Insectwingbeatsound Test & \textbf{34.20} \%&{\scriptsize\(\pm\) 8.1 \% } & 18.89 \%&{\scriptsize\(\pm\) 3.4 \% } \\
Italypowerdemand Test & \textbf{62.05} \%&{\scriptsize\(\pm\) 17.6 \% } & 49.85 \%&{\scriptsize\(\pm\) 0.0 \% } \\
Largekitchenappliances Test & 47.27 \%&{\scriptsize\(\pm\) 7.8 \% } & \textbf{57.73} \%&{\scriptsize\(\pm\) 0.2 \% } \\
Lightning2 Test & \textbf{54.10} \%&{\scriptsize\(\pm\) 0.0 \% } & \textbf{54.10} \%&{\scriptsize\(\pm\) 0.0 \% } \\
Lightning7 Test & \textbf{37.67} \%&{\scriptsize\(\pm\) 7.3 \% } & 36.99 \%&{\scriptsize\(\pm\) 5.8 \% } \\
Mallat Test & \textbf{15.63} \%&{\scriptsize\(\pm\) 6.6 \% } & 12.32 \%&{\scriptsize\(\pm\) 0.0 \% } \\
Meat Test & \textbf{67.50} \%&{\scriptsize\(\pm\) 1.7 \% } & 55.83 \%&{\scriptsize\(\pm\) 3.5 \% } \\
Medicalimages Test & \textbf{51.51} \%&{\scriptsize\(\pm\) 0.1 \% } & 51.45 \%&{\scriptsize\(\pm\) 0.0 \% } \\
Melbournepedestrian Test & 29.81 \%&{\scriptsize\(\pm\) 7.2 \% } & \textbf{62.63} \%&{\scriptsize\(\pm\) 0.8 \% } \\
Middlephalanxoutlineagegroup Test & \textbf{34.09} \%&{\scriptsize\(\pm\) 17.9 \% } & 18.83 \%&{\scriptsize\(\pm\) 0.0 \% } \\
Middlephalanxoutlinecorrect Test & \textbf{57.04} \%&{\scriptsize\(\pm\) 0.0 \% } & \textbf{57.04} \%&{\scriptsize\(\pm\) 0.0 \% } \\
Middlephalanxtw Test & \textbf{38.15} \%&{\scriptsize\(\pm\) 12.0 \% } & 27.27 \%&{\scriptsize\(\pm\) 0.0 \% } \\
Mixedshapesregulartrain Test & 48.98 \%&{\scriptsize\(\pm\) 14.4 \% } & \textbf{50.85} \%&{\scriptsize\(\pm\) 1.9 \% } \\
Mixedshapessmalltrain Test & 44.85 \%&{\scriptsize\(\pm\) 7.1 \% } & \textbf{49.20} \%&{\scriptsize\(\pm\) 1.0 \% } \\
Motestrain Test & \textbf{73.06} \%&{\scriptsize\(\pm\) 11.4 \% } & 71.45 \%&{\scriptsize\(\pm\) 8.3 \% } \\
Noninvasivefetalecgthorax1 Test & \textbf{06.78} \%&{\scriptsize\(\pm\) 9.9 \% } & 01.83 \%&{\scriptsize\(\pm\) 0.0 \% } \\
Noninvasivefetalecgthorax2 Test & \textbf{08.12} \%&{\scriptsize\(\pm\) 12.6 \% } & 02.29 \%&{\scriptsize\(\pm\) 0.6 \% } \\
Osuleaf Test & \textbf{29.65} \%&{\scriptsize\(\pm\) 8.1 \% } & 18.18 \%&{\scriptsize\(\pm\) 0.0 \% } \\
Oliveoil Test & \textbf{40.00} \%&{\scriptsize\(\pm\) 0.0 \% } & \textbf{40.00} \%&{\scriptsize\(\pm\) 0.0 \% } \\
Plaid Test & \textbf{23.32} \%&{\scriptsize\(\pm\) 6.1 \% } & 22.81 \%&{\scriptsize\(\pm\) 2.2 \% } \\
Phalangesoutlinescorrect Test & \textbf{61.31} \%&{\scriptsize\(\pm\) 0.0 \% } & \textbf{61.31} \%&{\scriptsize\(\pm\) 0.0 \% } \\
Phoneme Test & \textbf{13.13} \%&{\scriptsize\(\pm\) 2.0 \% } & 11.29 \%&{\scriptsize\(\pm\) 0.0 \% } \\
Pickupgesturewiimotez Test & 40.00 \%&{\scriptsize\(\pm\) 6.7 \% } & \textbf{54.00} \%&{\scriptsize\(\pm\) 2.8 \% } \\
Pigairwaypressure Test & 06.61 \%&{\scriptsize\(\pm\) 1.7 \% } & \textbf{19.71} \%&{\scriptsize\(\pm\) 0.0 \% } \\
Pigartpressure Test & 06.01 \%&{\scriptsize\(\pm\) 1.0 \% } & \textbf{25.24} \%&{\scriptsize\(\pm\) 1.0 \% } \\
Pigcvp Test & 10.58 \%&{\scriptsize\(\pm\) 8.7 \% } & \textbf{47.60} \%&{\scriptsize\(\pm\) 1.4 \% } \\
Plane Test & \textbf{26.67} \%&{\scriptsize\(\pm\) 16.7 \% } & 09.52 \%&{\scriptsize\(\pm\) 0.0 \% } \\
Powercons Test & \textbf{88.06} \%&{\scriptsize\(\pm\) 3.2 \% } & 85.00 \%&{\scriptsize\(\pm\) 0.8 \% } \\
Proximalphalanxoutlineagegroup Test & \textbf{66.22} \%&{\scriptsize\(\pm\) 19.9 \% } & 48.78 \%&{\scriptsize\(\pm\) 0.0 \% } \\
Proximalphalanxoutlinecorrect Test & \textbf{68.38} \%&{\scriptsize\(\pm\) 0.0 \% } & \textbf{68.38} \%&{\scriptsize\(\pm\) 0.0 \% } \\
Proximalphalanxtw Test & \textbf{50.49} \%&{\scriptsize\(\pm\) 17.8 \% } & 35.12 \%&{\scriptsize\(\pm\) 0.0 \% } \\
Refrigerationdevices Test & 41.20 \%&{\scriptsize\(\pm\) 4.5 \% } & \textbf{52.00} \%&{\scriptsize\(\pm\) 3.4 \% } \\
Rock Test & \textbf{44.50} \%&{\scriptsize\(\pm\) 3.4 \% } & 39.00 \%&{\scriptsize\(\pm\) 7.1 \% } \\ 
 \bottomrule
\end{tabular}
\end{table}

\begin{table}[h!]
\centering
\caption{UCR scores per dataset: S-Z}
\label{tab:classification-per-ds:s-z}
\begin{tabular}{@{}lrlrl@{}}
\toprule
\textbf{UCR Datasets}                & \multicolumn{2}{l}{\textbf{Our work}} & \multicolumn{2}{l}{\textbf{TS2Vec}}  \\ \midrule
Screentype Test & 37.33 \%&{\scriptsize\(\pm\) 5.4 \% } & \textbf{41.33} \%&{\scriptsize\(\pm\) 0.8 \% } \\
Semghandgenderch2 Test & \textbf{68.29} \%&{\scriptsize\(\pm\) 5.8 \% } & 55.25 \%&{\scriptsize\(\pm\) 5.1 \% } \\
Semghandmovementch2 Test & 27.44 \%&{\scriptsize\(\pm\) 7.0 \% } & \textbf{32.00} \%&{\scriptsize\(\pm\) 4.4 \% } \\
Semghandsubjectch2 Test & \textbf{40.67} \%&{\scriptsize\(\pm\) 8.7 \% } & 33.67 \%&{\scriptsize\(\pm\) 2.7 \% } \\
Shakegesturewiimotez Test & 46.00 \%&{\scriptsize\(\pm\) 6.3 \% } & \textbf{71.00} \%&{\scriptsize\(\pm\) 9.9 \% } \\
Shapeletsim Test & 60.83 \%&{\scriptsize\(\pm\) 8.7 \% } & \textbf{83.33} \%&{\scriptsize\(\pm\) 6.3 \% } \\
Shapesall Test & 29.46 \%&{\scriptsize\(\pm\) 1.0 \% } & \textbf{43.00} \%&{\scriptsize\(\pm\) 0.5 \% } \\
Smallkitchenappliances Test & 48.60 \%&{\scriptsize\(\pm\) 9.5 \% } & \textbf{55.33} \%&{\scriptsize\(\pm\) 0.2 \% } \\
Smoothsubspace Test & 61.17 \%&{\scriptsize\(\pm\) 5.3 \% } & \textbf{77.33} \%&{\scriptsize\(\pm\) 0.9 \% } \\
Sonyaiborobotsurface1 Test & \textbf{47.67} \%&{\scriptsize\(\pm\) 8.1 \% } & 42.93 \%&{\scriptsize\(\pm\) 0.0 \% } \\
Sonyaiborobotsurface2 Test & \textbf{67.24} \%&{\scriptsize\(\pm\) 7.7 \% } & 61.70 \%&{\scriptsize\(\pm\) 0.0 \% } \\
Starlightcurves Test & 76.83 \%&{\scriptsize\(\pm\) 12.8 \% } & \textbf{83.96} \%&{\scriptsize\(\pm\) 1.1 \% } \\
Strawberry Test & \textbf{64.32} \%&{\scriptsize\(\pm\) 0.0 \% } & \textbf{64.32} \%&{\scriptsize\(\pm\) 0.0 \% } \\
Swedishleaf Test & \textbf{25.72} \%&{\scriptsize\(\pm\) 6.0 \% } & 07.84 \%&{\scriptsize\(\pm\) 0.9 \% } \\
Symbols Test & \textbf{43.52} \%&{\scriptsize\(\pm\) 20.0 \% } & 17.39 \%&{\scriptsize\(\pm\) 0.0 \% } \\
Syntheticcontrol Test & 62.67 \%&{\scriptsize\(\pm\) 14.2 \% } & \textbf{82.17} \%&{\scriptsize\(\pm\) 6.8 \% } \\
Toesegmentation1 Test & 62.17 \%&{\scriptsize\(\pm\) 5.9 \% } & \textbf{66.89} \%&{\scriptsize\(\pm\) 1.6 \% } \\
Toesegmentation2 Test & \textbf{75.77} \%&{\scriptsize\(\pm\) 11.8 \% } & 73.08 \%&{\scriptsize\(\pm\) 10.9 \% } \\
Trace Test & \textbf{52.25} \%&{\scriptsize\(\pm\) 11.1 \% } & 19.00 \%&{\scriptsize\(\pm\) 0.0 \% } \\
Twoleadecg Test & \textbf{51.51} \%&{\scriptsize\(\pm\) 3.1 \% } & 49.96 \%&{\scriptsize\(\pm\) 0.0 \% } \\
Twopatterns Test & \textbf{34.88} \%&{\scriptsize\(\pm\) 4.1 \% } & 25.87 \%&{\scriptsize\(\pm\) 0.0 \% } \\
Umd Test & 52.78 \%&{\scriptsize\(\pm\) 5.9 \% } & \textbf{55.90} \%&{\scriptsize\(\pm\) 8.3 \% } \\
Uwavegesturelibraryall Test & \textbf{26.95} \%&{\scriptsize\(\pm\) 18.5 \% } & 12.33 \%&{\scriptsize\(\pm\) 0.1 \% } \\
Uwavegesturelibraryx Test & \textbf{57.49} \%&{\scriptsize\(\pm\) 4.6 \% } & 17.23 \%&{\scriptsize\(\pm\) 7.1 \% } \\
Uwavegesturelibraryy Test & \textbf{48.75} \%&{\scriptsize\(\pm\) 6.1 \% } & 19.11 \%&{\scriptsize\(\pm\) 1.8 \% } \\
Uwavegesturelibraryz Test & \textbf{51.70} \%&{\scriptsize\(\pm\) 5.4 \% } & 25.49 \%&{\scriptsize\(\pm\) 8.3 \% } \\
Wafer Test & \textbf{90.97} \%&{\scriptsize\(\pm\) 3.5 \% } & 89.21 \%&{\scriptsize\(\pm\) 0.0 \% } \\
Wine Test & \textbf{50.00} \%&{\scriptsize\(\pm\) 0.0 \% } & \textbf{50.00} \%&{\scriptsize\(\pm\) 0.0 \% } \\
Wordsynonyms Test & \textbf{27.35} \%&{\scriptsize\(\pm\) 4.5 \% } & 21.94 \%&{\scriptsize\(\pm\) 0.0 \% } \\
Wormstwoclass Test & \textbf{58.44} \%&{\scriptsize\(\pm\) 2.6 \% } & 57.14 \%&{\scriptsize\(\pm\) 0.0 \% } \\
Worms Test & \textbf{43.83} \%&{\scriptsize\(\pm\) 1.9 \% } & 42.86 \%&{\scriptsize\(\pm\) 0.0 \% } \\
Yoga Test & 53.34 \%&{\scriptsize\(\pm\) 0.3 \% } & \textbf{53.57} \%&{\scriptsize\(\pm\) 0.0 \% } \\
\bottomrule
\end{tabular}
\end{table}

\FloatBarrier
\textit{End of the Appendix.}

\end{document}